\documentclass[10pt,twocolumn,letterpaper]{article}

\usepackage[pagenumbers]{wacv} % To force page numbers, e.g. for an arXiv version

\usepackage{graphicx}
\usepackage{amsmath}
\usepackage{amssymb}
\usepackage{booktabs}
\usepackage{enumitem}
\usepackage{dblfloatfix}
\usepackage{balance}

\usepackage[pagebackref,breaklinks,colorlinks]{hyperref}

\usepackage[capitalize]{cleveref}
\crefname{section}{Sec.}{Secs.}
\Crefname{section}{Section}{Sections}
\Crefname{table}{Table}{Tables}
\crefname{table}{Tab.}{Tabs.}

\def\wacvPaperID{2362} % *** Enter the WACV Paper ID here
\def\confName{WACV}
\def\confYear{2024}

\usepackage{fancyhdr}
\fancypagestyle{firstpage}{
  \fancyhf{} % clear all header and footer fields
  \fancyhead[C]{To appear in proceedings of the 2024 IEEE/CVF Winter Conference on Applications of Computer Vision (WACV) \\ The final publication will be available soon.} % set the header
  \headsep = 0.7cm
  \fancyfoot[C]{1}
}

\begin{document}

%%%%%%%%% TITLE - PLEASE UPDATE
\title{Joint 3D Shape and Motion Estimation from Rolling Shutter Light-Field Images}

\author{
%\phantom{000000000} 
Hermes McGriff$^{~1,3}$ \hspace{0.2in}
Renato Martins$^{1,2}$   \hspace{0.2in}
Nicolas Andreff$^{~3}$   \hspace{0.2in}
Cédric Demonceaux$^{1,2}$ \vspace{0.05in}\\
$^1$Université de Bourgogne, CNRS UMR 6303 ICB \hspace{0.04in}
$^2$Université de Lorraine, CNRS, Inria, LORIA \hspace{0.04in}\\
$^3$Université de Franche-Comté, CNRS UMR 6174 FEMTO-ST\\
{\tt\footnotesize \{hermes.mc-griff,renato.martins,cedric.demonceaux\}@u-bourgogne.fr, nicolas.andreff@univ-fcomte.fr}
}

\maketitle

\thispagestyle{firstpage}
%%%%%%%%% ABSTRACT
\begin{abstract}
In this paper, we propose an approach to address the problem of 3D reconstruction of scenes from a single image captured by a light-field camera equipped with a rolling shutter sensor. Our method leverages the 3D information cues present in the light-field and the motion information provided by the rolling shutter effect. We present a generic model for the imaging process of this sensor and a two-stage algorithm that minimizes the re-projection error while considering the position and motion of the camera in a motion-shape bundle adjustment estimation strategy. Thereby, we provide an instantaneous 3D shape-and-pose-and-velocity sensing paradigm. To the best of our knowledge, this is the first study to leverage this type of sensor for this purpose. We also present a new benchmark dataset composed of different light-fields showing rolling shutter effects, which can be used as a common base to improve the evaluation and tracking the progress in the field. We demonstrate the effectiveness and advantages of our approach through several experiments conducted for different scenes and types of motions. The source code and dataset are publicly available at: \url{https://github.com/ICB-Vision-AI/RSLF}.
\end{abstract}

%%%%%%%%% BODY TEXT
\section{Introduction}
Light-field (LF) cameras (also known as plenoptic), introduced by Adelson and Wang~\cite{adelson1992single} and prototyped by Ng~\cite{ng2005light}, consist of a conventional camera with a micro-lens array in front of the photosensitive sensor. This type of imaging sensor has the particularity of being able to capture a light field of a scene in a single capture. LF cameras are now an established solution used in computer vision, photogrammetry and robotics \cite{conti2020dense, ihrke2016principles, wu2017light}.
\begin{figure}[t]
	\centering	
	\hspace*{-0.23cm}\includegraphics[width=1.05\linewidth]{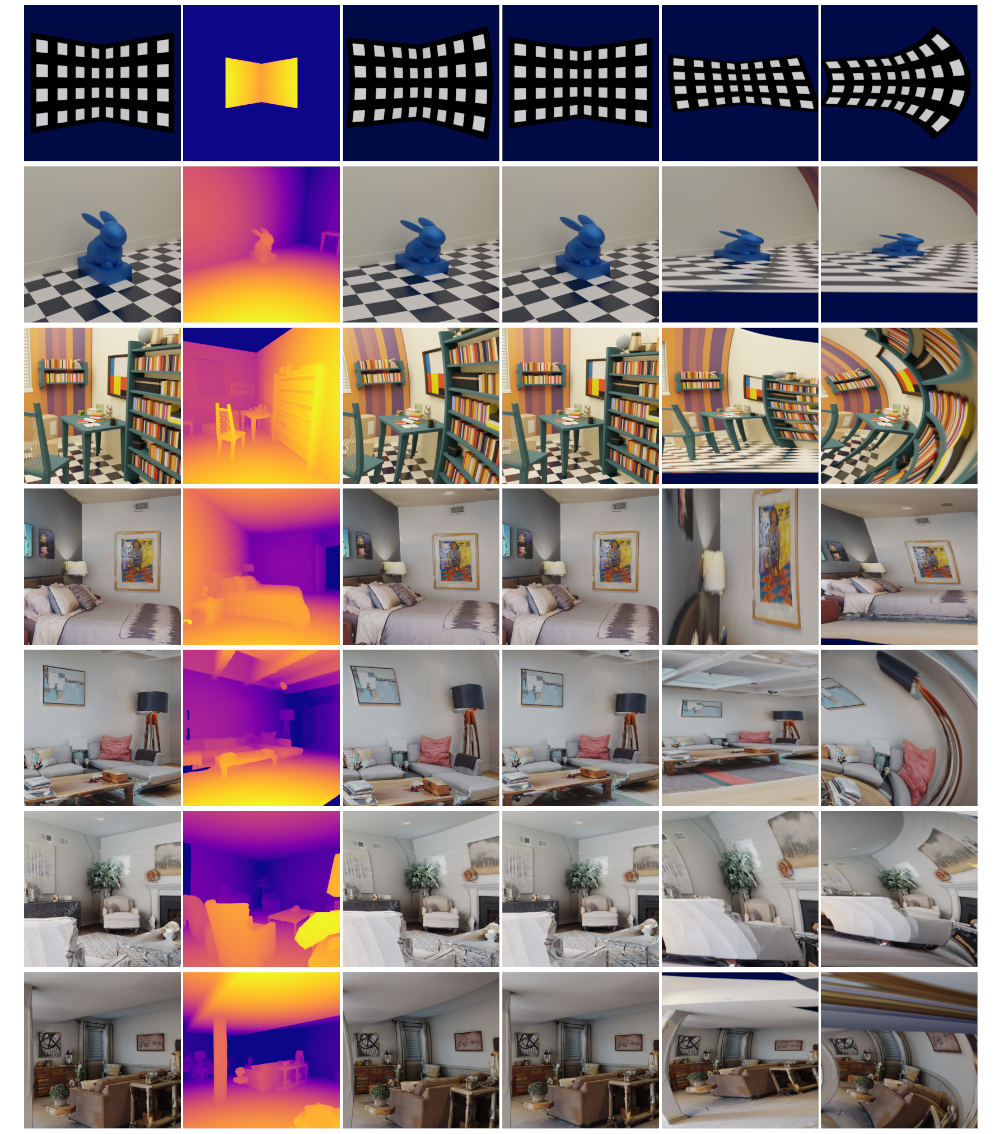}
	\caption{Some central views of the proposed RSLF dataset. From left to right: A global shutter view, the ground truth depth, and four different rolling shutter views with increasing camera motions.}
	\label{fig:simu}
\end{figure}
The miniaturization of these cameras, \eg in the context of applications such as intra-corporeal micro-robotics, requires the choice of a rolling shutter (RS) photosensitive sensor. Conversely to global shutter (GS) cameras, where all pixels of the image are acquired at the same time, the image acquisition by rolling shutter is sequential \cite{meingast2005geometric}. Notably, a RS sensor creates image deformations in the case of dynamic scenes or when the camera is moving, as depicted in \cref{fig:simu}. The rolling shutter then often degrades the performance and challenge existing reconstruction and pose estimation approaches. Ait-Aider \etal~\cite{ait2006simultaneous} have shown that, in the case of a conventional perspective monocular camera, these deformations can be leveraged in order to compute the motion of the scene with respect to the camera. However, their proposed model and subsequent improved strategies~\cite{ait2009structure, LAO20181} have the strong limitation of requiring the shape of the object/scenes to be known. Conversely, this paper proposes to jointly estimate the motion and the structure of a scene from a single view shot and in less constrained conditions. Although the possibilities given by RS, when properly modelled, has been shown for several computer vision and graphics problems, the combination with LF has not been yet exploited in a unified approach. One important motivation of this paper is to show the possibilities that this sensor modality presents, such as of being able to allow the estimation of the camera motion (or from the scene/object) from a single view without priors on the scene shape.
We are particularly motivated by showing the interest of a unified approach (and its properties) that is capable of leveraging RS with existing light-field consumer devices. Indeed, such sensors are available, like the entry level cameras of Raytrix (R8, R42, R10$\mu$, R20) or any camera array with RS sensors (like Pelican Imaging), but unfortunately no public dataset is available to be the best of the authors' knowledge. In this context, another core motivation of this paper is to present a suitable and challenging LF dataset collected with a RS camera with different motion levels and scene geometries. For that, we have generated new models and leveraged existing scene models (from Matterport3D) into an adapted rendering engine (based on Blender) to create LF data affected by RS distortions in different conditions (e.g., from mild to strong motions). The main contributions of this paper are as follows:
\begin{itemize}
	\item We propose a generic projection model of a rolling shutter light-field (RSLF) camera. This model is capable to represent a light-field with both global shutter and rolling shutter settings. \vspace{-0.3cm}
	\item A non-linear bundle adjustment strategy is designed to estimate jointly the 3D shape and motion for this sensor modality. We also design a linear initialization strategy in order to recover a first coarse estimate of the 3D shape. This initialization is essential for the convergence of our approach as shown in the ablation studies.\vspace{-0.3cm}
    \item We also present a new dataset composed of Rolling Shutter Light Fields (RSLF) paired with ground truth depth maps, on several synthetic scenes and with different types and levels of motion. We aim this dataset to be used as a common base to improve the evaluation and help tracking the progress in the field.
\end{itemize}

%-------------------------------------------------------------------------
\section{Related Work}
\paragraph{Depth estimation from light-fields.} Light-field contains rich information cues about the geometry of the scene. The seminal work of Adelson and Wang~\cite{adelson1992single} for the plenoptic camera exploit this ability for ``single lens stereo''. They used sub-aperture images (SAI) to perform a standard two frame displacement analysis with multiple pairs horizontally and vertically. In the same direction, multi-view stereo matching-based methods try to reproduce the results of classical stereo with plenoptic images~\cite{georgiev2010focused,perwass2012single,jeon2015accurate,zeller2016depth}. In this context, Georgiev and Lumsdaine~\cite{georgiev2010focused} introduced the focused plenoptic camera and proposed a complete setup in order to recover depth with this new design. The method simultaneously render the image and estimate a per micro-lens depth map by computing the cross correlation between patches in micro images. Similarly, Perwass and Wietzke~\cite{perwass2012single} introduced a multi-focused plenoptic camera model alongside a depth estimation algorithm based on point correspondences between micro images. Jeon and Park~\cite{jeon2015accurate} explored the phase-shift theorem of the Fourier transform to estimate an accurate sub-pixel disparity map by computing a matching cost volume between SAI. Zeller \etal~\cite{zeller2016depth} proposed a filtering method for the estimation of semi-dense probabilistic depth maps for focused plenoptic cameras, with a Kalman filter like approach preserving discontinuities in the depth map. Ferreira and Goncalves ~\cite{ferreira2016fast} proposed a similar but faster depth map estimation method, with SIFT correspondences and through epipolar lines on the micro images. Bok \etal ~\cite{bok2016geometric} proposed a calibration of the light-field camera based on a bundle adjustment method and Zhang \etal~\cite{zhang2018generic} proposed a generic multi-projection model (along with its calibration algorithm) for LF cameras. Most of these techniques rely on generating SAI and then applying classic stereo matching algorithms to estimate the depth of the scene. However, they assume GS cameras (or with slow moving objects and camera motions). Our approach, on the other hand, can handle scenes with a camera in movement and is far less affected by RS distortions due to camera motions.\vspace{-0.3cm}

\paragraph{Epipolar plane images and learning-based LF analysis.} The scene structure can also be extracted from Epipolar Plane Images (EPI)~\cite{bolles1987epipolar, dansereau2004gradient,wanner2012globally, NEURIPS2021_a11ce019}. These approaches estimate depth information from the slopes of the lines observed in the Epipolar planes. Tao et al.~\cite{tao2015depth} improved the accuracy of the depth estimation with a weighted sum between the defocus and correspondence cues present in EPIs. Zhang \etal ~\cite{zhang2016robust} proposed a spinning parallelogram operator to determine the line slopes. Lin \etal~\cite{lin2015depth} leveraged the refocus capability of light-fields and the possibility to use Shape-From-Focus. Closely related to our work, Srinivasan \etal~ \cite{srinivasan2017light} proposed a motion estimation from a single view with a light-field camera based on motion blur. Heber and Pock~\cite{heber2016convolutional} first used a Convolutional Neural Network to compute depth from light-field images. Shin \etal~ \cite{shin2018epinet} proposed a fast and accurate light field depth estimation method based on a fully-convolutional neural network and a light-field image-specific data augmentation. These techniques suffer by the lack of generalization to new/unseen scenes and often dependence on significant amount of data.\vspace{-0.3cm}

\paragraph{Rolling shutter structure-from-motion estimation.} The potential of RS images received increased attention for scene analysis. Meingast \etal~\cite{meingast2005geometric} developed a general projection equation for a rolling shutter camera and also proposed a calibration to estimate the rate of the rolling shutter. Ait-Aider \etal~ \cite{ait2006simultaneous} first showed that the rolling shutter effect could be leveraged to estimate the motion of an object, but of known shape, when the majority of previous studies on the rolling shutter were about compensating it~\cite{liang2008analysis,karpenko2011digital}. This is notably done for blur compensation with both model and learning-based approaches~\cite{meilland2013unified,fan2021sunet}. Saurer \etal~ \cite{saurer2013rolling} and Ait-Aider \etal~ \cite{ait2009structure} investigated RS effects for stereo vision. Recently Lao \etal~ \cite{lao2021solving} proposed a analogy with non-rigidity to solve shape estimation with a monocular rolling shutter image. Different than these previous methods, we address the ambiguity between shape and motion inherent to RS images exploiting the properties of the LF. We show that a micro-lens array in front of the RS sensor allows to model the RS effect in the case of 3D scenes and to estimate the movement of the scene with respect to the camera without prior knowledge of the scene geometry.

\section{Method}
Our joint 3D scene reconstruction and camera motion estimation approach has two main stages. Firstly, a coarse linear solution is computed to provide an initialization for a non-linear bundle adjustment method. This method is designed to handle the geometric and temporal constraints that are present in the Rolling Shutter Light-Field setting. \vspace*{-0.3cm}
\paragraph{Light-field modeling and RS projection.}
\begin{figure*}[t!]
        \centering
	    \includegraphics[height=0.27\linewidth]{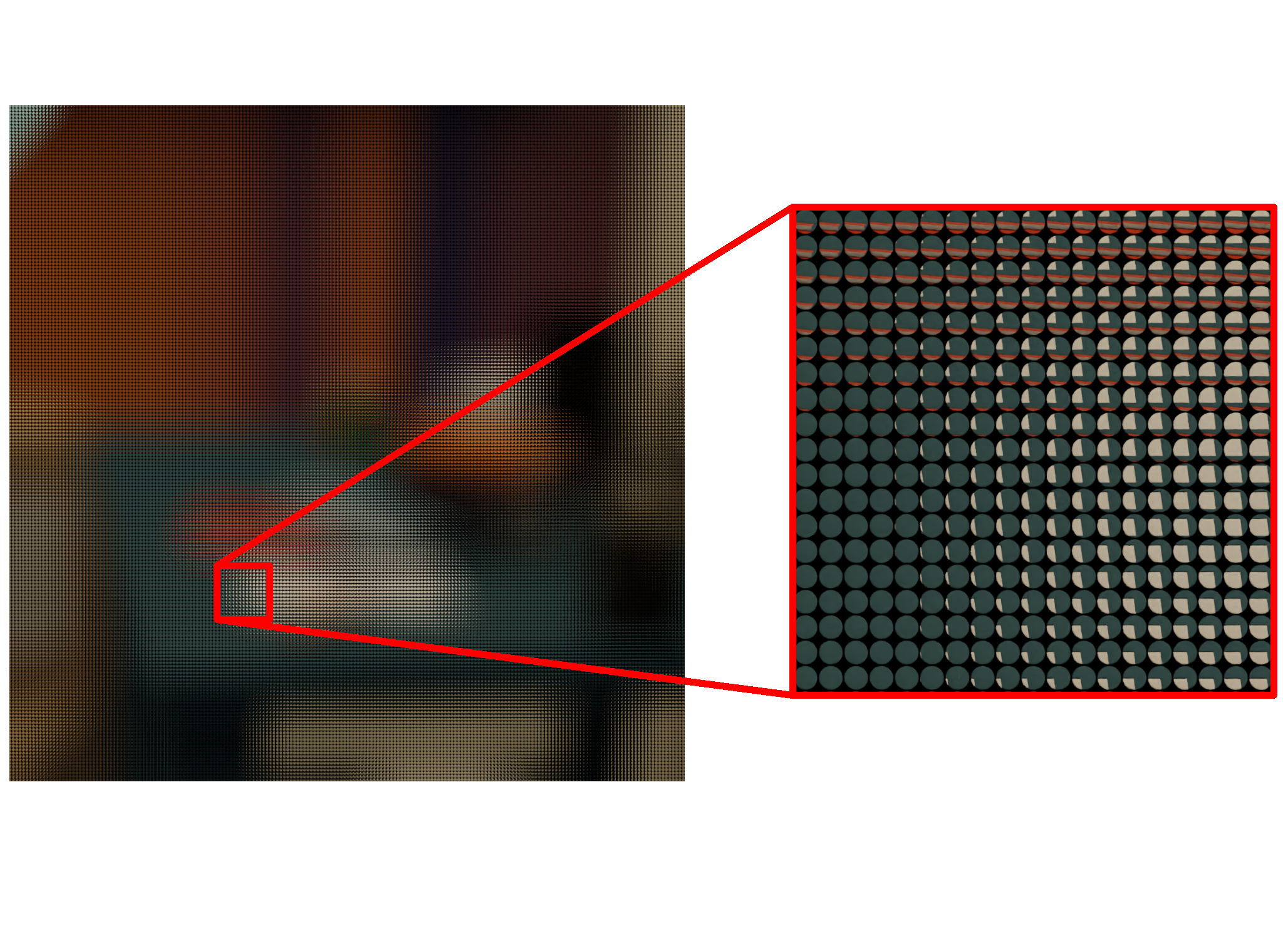}	\hspace*{0.3cm}
	    \includegraphics[height=0.33\linewidth]{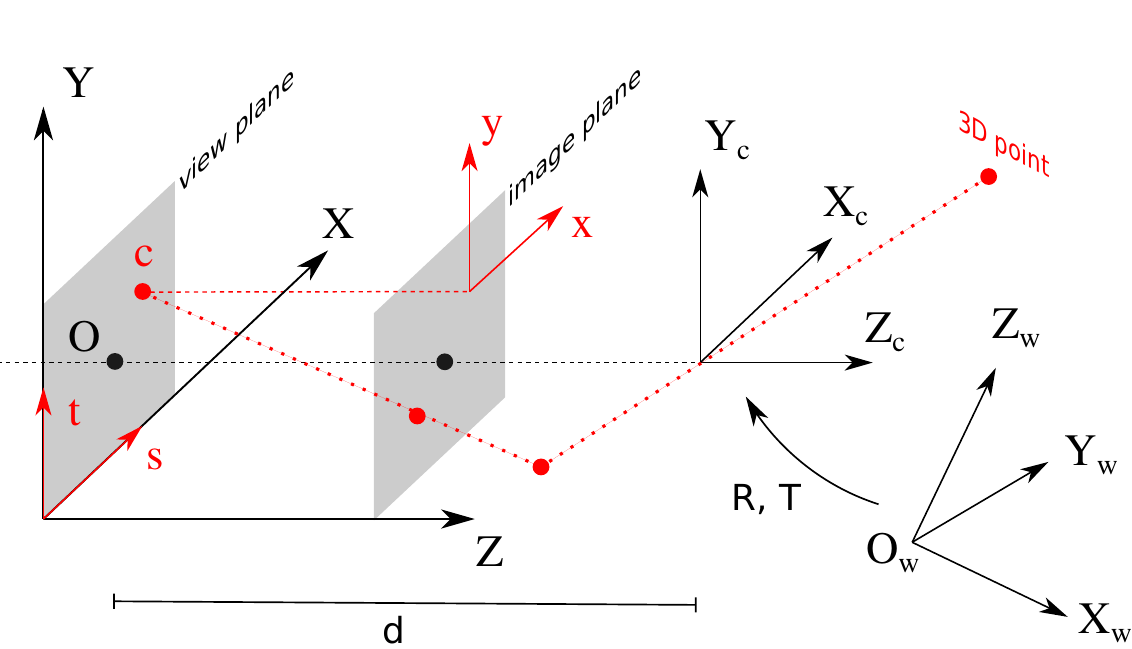}	
	    \caption{\textit{left} - A raw plenoptic image from a near viewpoint in the scene shown in \cref{fig:simu} and a detail of the micro-images. \textit{right} - The adopted LF coordinate frames: The 3D point is projected in a 3D virtual scene by thin lens projection, then on the 2D image plane by pinhole projection which coordinate frame depends on the considered viewpoint.}\vspace*{-0.3cm}
	    \label{fig:frames}
\end{figure*}
To provide a comprehensive theoretical framework for our proposed approach, we begin by presenting an overview of the light field projection modeling. Subsequently, we use this framework to introduce a projection model formulation that is specifically designed for the RSLF setting. A more detailed description of the projection model formulation and theoretical analysis are given in the Supplementary material.
An overview of the adopted light-field modeling and geometry is shown in \cref{fig:frames}. The pose of the camera with respect to the scene expressed in the world coordinates frame $(\mathbf{O}_w, X_w, Y_w, Z_w)$ is $[\mathbf{R}\mid\mathbf{T}] \in \mathbb{SE}(3)$. The camera position defines a new coordinate frame $(\mathbf{O}_c, X_c, Y_c, Z_c)$ with origin  placed in the optical center of the main lens. The view plane (MLA plane) has coordinate frame $(\mathbf{O}, X, Y, Z)$ expressed in relation to the camera frame by a pure translation $(O_x, O_y, d)$ expressed by the transformation matrix $\mathbf{D} \in SE(3)$, with $\mathbf{O}=(O_x, O_y, 0)^\top$ the intersection of the optical axis and the view plane, and $d$ the distance between the optical center of the main lens and the view plane. % related to the geometry of the LF camera.
The micro-image local frames $(x, y)$ are attached to the image plane and are dependent of the considered viewpoint, as shown in \cref{fig:frames}. Given a point in the world homogeneous coordinates frame ${}^w \widetilde{\mathbf{p}} =(x_w, y_w, z_w, 1)^\top$ and the matrices ${}^c \mathbf{M}_w$ (the transformation between the camera to world coordinates) and  $\mathbf{K}_c$ (thin lens projection matrix), we can obtain the virtual projection of the 3D point inside the camera as:
\begin{equation}
\lambda_c
\widetilde{\mathbf{p}} = \mathbf{D}\mathbf{K}_c {}^c \mathbf{M}_w {}^w \widetilde{\mathbf{p}},
\label{eq:proj_w}
\end{equation}
with $\lambda_c$ a scaling factor. For a given viewpoint $\mathbf{c}=(s,t,0)^\top$, \ie a projection center, the projection of the point $\widetilde{\mathbf{p}}$ onto the image plane is given by:
\begin{equation}
\label{eq:mat_zhang}
\lambda_s^{s,t}
\widetilde{\mathbf{m}}^{s,t} = \mathbf{K}_{s}^{s,t}\widetilde{\mathbf{p}}
=	\begin{bmatrix}
	f & 0 & 0 & -fs\\
	0 & f & 0 & -ft\\
	0 & 0 & 1 & 0\\
	\end{bmatrix}
\widetilde{\mathbf{p}},
\end{equation}
with $\widetilde{\mathbf{m}}^{s,t} =(x^{s,t}, y^{s,t}, 1)^\top$ the final LF image points, $f$ the focal length of the micro-lenses and $\lambda_s$ a scaling factor.

\paragraph{Rolling shutter modeling.} We follow a similar formalism to Ait-Aider \etal~\cite{ait2006simultaneous} to represent an RS imaging process. The main insight is to define a projection model dependent of the camera pose and as a function of the micro-image line $t$ being observed. We adopt the hypothesis that the speeds $(\mathbf{v},\Omega)$ are constant during the LF acquisition. Adapting the initial projection defined in \cref{eq:proj_w} for the RS we have:
\begin{equation}
\label{eq:deltaM}
\lambda_c
\widetilde{\mathbf{p}} = \mathbf{D}\mathbf{K}_c
\left[\begin{array}{c c} 
\delta\mathbf{R}^t {}^c \mathbf{R}_w & {}^c \mathbf{T}_w +\delta\mathbf{T}^t\\ 
\boldsymbol{0}^\top & 1 
\end{array}\right]
{}^w \widetilde{\mathbf{p}},
\end{equation}
with $\delta\mathbf{R}^t = \mathbf{a}\mathbf{a}^\top (1-\cos(\Omega \tau t))+\mathbf{I} \cos(\Omega \tau t)+[\mathbf{a}]_\wedge \sin(\Omega \tau t), \mbox{ and } \delta\mathbf{T}^t = \mathbf{v} \tau t,$
where $\mathbf{a}$ (axis of rotation), $\Omega$ (angular velocity) and $\mathbf{v}$ (linear velocity) describe the uniform movement of the camera coordinate frame with respect to the world coordinates frame and $\tau$ the time between the acquisition of two lines of the micro-images. The full Rolling Shutter LF projection from \cref{eq:deltaM} that projects the 3D point ${}^w \widetilde{\mathbf{p}}_i$ to an image point $\mathbf{m}_i^{s,t} \in \mathbb{P}^2$, given a center of projection $\mathbf{c}=(s,t,0)^\top$ is then
\begin{equation}
\label{eq:proj}
\lambda \mathbf{m}_i^{s,t} = \mathbf{K}_s^{s,t} \mathbf{D} \mathbf{K}_c [\delta\mathbf{R}^t {}^c \mathbf{R}_w \mid {}^c \mathbf{T}_w +\delta\mathbf{T}^t] {}^w \widetilde{\mathbf{p}}_i,
\end{equation}
where $\mathbf{K}_s^{s,t} \mathbf{D} \mathbf{K}_c$ can be represented as a single compact intrinsic Rolling Shutter LF tensor:
\begin{equation}
\label{eq:Kst}
\mathbf{K}_s^{s,t} \mathbf{D} \mathbf{K}_c =
\begin{bmatrix}
f & 0 & -\frac{f}{F}(O_x-s) & f(O_x-s)\\
0 & f & -\frac{f}{F}(O_y-t) & f(O_y-t)\\
0 & 0 & 1-\frac{d}{F} & d
\end{bmatrix},
\end{equation}
\noindent with $F$ the focal length of the main lens. This formulation has the strong advantage of being generic and represent both GS and RS configurations. Another advantage is that all parameters of this unified model can be calibrated with existing techniques such as Bok \etal \cite{bok2016geometric} for the intrinsic parameters and Meingast \etal \cite{meingast2005geometric} for the rolling shutter rate.

\paragraph{Generalization and particular cases.} When $\tau = 0$ (\ie, no temporal delay between two consecutive lines), this model can be simplified to a GS light-field camera as the position of the sensor with respect to the scene will be identical for any $t$. The situation where the camera has no velocity with respect to the scene can also be seen as GS for similar reasons. The proposed projection model in \cref{eq:proj} generalizes to a conventional pinhole camera projection in the case where the MLA is composed of a unique lens. 
More details are given in the Supplementary material.

\subsection{Scene Structure and Motion Estimation}
For a given set of matching points inside a calibrated LF and assuming that all points belongs to the same rigid scene in a uniform movement with respect to the camera, we can recover the position of the points in the 3D world as well as their motion at a given time. We will use a re-projection error minimization in order to find jointly these 3D coordinates ${}^w \widetilde{\mathbf{p}}_i$ and the dynamic parameters describing the movement of the camera.
\paragraph{Linear initialization.}
A classical multi-view stereo strategy is applied to provide a first estimate of the 3D points in the scene. In order to reduce the influence of the RS effect, we apply the multi-view stereo only horizontally, thereby ensuring that each measured point ${}^w \widetilde{\mathbf{p}}_i \in \mathbb{R}^3$ is captured at the same instant. From the experiments, this first estimate is essential to allow convergence of the following non-linear optimization.
\paragraph{Non-linear bundle adjustment.}
Using this 3D initialization of the observed points in the light field and our projection model, we design a re-projection error in order to recover simultaneously a refined structure of the scene and the camera motion. From our projection in \cref{eq:proj} we compute the point:
\begin{equation}
\left(
u_i^{s,t},
v_i^{s,t},
w_i^{s,t}
\right)^T
= \mathbf{K}_s^{s,t} \mathbf{D} \mathbf{K}_c [\delta\mathbf{R}^t \mid \delta\mathbf{T}^t] {}^w \widetilde{\mathbf{p}}_i
\end{equation}
and deduce the Euclidean pixel coordinates, the scalars $x_i^{s,t}$ and $y_i^{s,t}$, computed as:
\begin{equation}
\label{eq:reproj}
\begin{aligned}
x_i^{s,t} &= \frac{u_i^{s,t}}{w_i^{s,t}} := \xi_{(x)}^{s,t} ({}^w \widetilde{\mathbf{p}}_i, \Omega, \mathbf{a}, \mathbf{v}), \text{\hspace*{0.5cm} and} \\ 
y_i^{s,t} &= \frac{v_i^{s,t}}{w_i^{s,t}} := \xi_{(y)}^{s,t} ({}^w \widetilde{\mathbf{p}}_i, \Omega, \mathbf{a}, \mathbf{v}),
\end{aligned}
\end{equation}
\noindent with $\xi^{s,t}$ the projection function that, given a center of projection $\mathbf{c}=(s,t,0)^\top$, return the coordinates of the image point with respect to its static position and its movement. The re-projection error function is obtained by computing the distance between the measured points $\mathbf{\Tilde{m}}_i^{s, t_0}(\Tilde{x_i}^{s,t}, \Tilde{y_i}^{s,t})$ and the coordinates estimated with $ \xi_{(x)}^{s,t}$ and $\xi_{(y)}^{s,t}$ from \cref{eq:reproj} as follows:
\begin{equation}
\label{eq:error_func}
\begin{aligned}
    \epsilon = &\sum_{s} \sum_{t} \sum_{i} \left(  \Tilde{x_i}^{s,t} - \xi_{(x)}^{s,t} ({}^w \widetilde{\mathbf{p}}_i, \Omega, \mathbf{a}, \mathbf{v}) \right)^2 \\ &+ \left(  \Tilde{y_i}^{s,t} - \xi_{(y)}^{s,t} ({}^w \widetilde{\mathbf{p}}_i, \Omega, \mathbf{a}, \mathbf{v}) \right)^2.
\end{aligned}
\end{equation}

\noindent This problem has three unknowns for $\Omega \mathbf{a}$, three unknowns for $\mathbf{v}$, and three unknowns for every ${}^w \widetilde{\mathbf{p}}_i$. It can be solved if at least four non-coplanar 3D points can be observed, meaning that they need to be located at least an LF image in two different lines and at two different columns of micro-images.

\paragraph{Regularization.}
For the moment, the rotation axis $\mathbf{a}$ in \cref{eq:error_func} is defined to pass through the world origin, which corresponds to the optical center of the main lens. However, this is generally not the instantaneous center of rotation of the movement between the camera and the scene. To ease the description of the movement, we regularize the optimization by providing a ``center of rotation'' $\mathbf{g}$ to the point cloud. This allows to express all points ${}^w\mathbf{p}_i$ in a new coordinate frame centered on this center of rotation. It also allows to compute normalized points ${}^n\mathbf{p}_i$ from which the coordinates are lying in the range $[-1, 1]$. The final non-linear adapted re-projection error from \cref{eq:error_func} using the normalized points and the center of rotation regularization is then:
\begin{equation}
\label{eq:error_func2}
\begin{aligned}
    \epsilon = &\sum_{s} \sum_{t} \sum_{i} \left(  \Tilde{x_i}^{s,t} - {}^n\xi_{(x)}^{s,t} ({}^n \mathbf{p}_{i}, \mathbf{g}, \Omega, \mathbf{a}, \mathbf{v}) \right)^2 \\ &+ \left(  \Tilde{y_i}^{s,t} - {}^n\xi_{(y)}^{s,t} ({}^n \mathbf{p}_{i}, \mathbf{g},\Omega, \mathbf{a}, \mathbf{v}) \right)^2,
\end{aligned}
\end{equation}
\noindent where ${}^n\xi_{(x)}^{s,t}$ and ${}^n\xi_{(y)}^{s,t}$ are designed to handle the normalization, and $\mathbf{g}$ is also optimized in the loop so that the model is able to find the optimal center of rotation on-the-fly. Further details on the optimization are provided in the supplementary materials.

\section{Rolling Shutter Light-Field Dataset}

\begin{table*}[t!]
	\centering
	\resizebox{1.3\columnwidth}{!}{%
		\begin{tabular}{llllllllll}
			\toprule
			& \multicolumn{3}{c}{abs rel $\downarrow$} & \multicolumn{3}{c}{abs diff $\downarrow$} & \multicolumn{3}{c}{RMS $\downarrow$}\\
			\cmidrule(lr){2-4} \cmidrule(lr){5-7} \cmidrule(lr){8-10}
            Method & GS & slow & fast & GS & slow & fast& GS & slow & fast\\
        	\cmidrule(lr){1-1} \cmidrule(lr){2-4} \cmidrule(lr){5-7} \cmidrule(lr){8-10}
Jeon-CVPR \cite{jeon2015accurate} &      \textbf{0.040} &      \underline{0.053} &      \underline{0.110} &      \textbf{0.027} &      \underline{0.036} &      \underline{0.071} &      \textbf{0.035} &      \textbf{0.048} &      \underline{0.092} \\
OACC-Net \cite{OACC-Net} &      0.143 &      0.171 &      0.196 &      0.091 &      0.109 &      0.125 &      0.109 &      0.128 &      0.144 \\
Ours &      \textbf{0.040} &      \textbf{0.041} &      \textbf{0.059} &      \underline{0.031} &      \textbf{0.032} &      \textbf{0.044} &      \underline{0.046} &      \underline{0.051} &      \textbf{0.064} \\

            \bottomrule
            \toprule
            & \multicolumn{3}{c}{$\delta < 1.25 \uparrow$} & \multicolumn{3}{c}{$\delta < 1.25^2 \uparrow$} & \multicolumn{3}{c}{$\delta < 1.25^3 \uparrow$}\\
            \cmidrule(lr){2-4} \cmidrule(lr){5-7} \cmidrule(lr){8-10}
            Method & GS & slow & fast& GS & slow & fast& GS & slow & fast\\
            \cmidrule(lr){1-1} \cmidrule(lr){2-4} \cmidrule(lr){5-7} \cmidrule(lr){8-10}
Jeon-CVPR \cite{jeon2015accurate} &      \textbf{0.993} &      \textbf{0.976} &      \underline{0.894} &      \textbf{1.000} &      \textbf{0.999} &      \underline{0.973} &      \textbf{1.000} &      \textbf{1.000} &      \underline{0.998} \\
OACC-Net \cite{OACC-Net} &      0.767 &      0.720 &      0.676 &      0.959 &      0.945 &      0.933 &      \textbf{1.000} &      0.997 &      0.997 \\
Ours &      \underline{0.958} &      \underline{0.961} &      \textbf{0.949} &      \underline{0.989} &      \underline{0.988} &      \textbf{0.982} &      0.999 &      \underline{0.999} &     \textbf{0.999} \\

             \bottomrule
		\end{tabular}
  }
 \vspace*{-0.2cm}
 \caption{Average reconstruction error metrics in different scenarios for all dataset sequences: \textit{GS} (global shutter, equivalent to a static camera scenario), \textit{slow} (RS with small camera linear and angular velocities), and \textit{fast} (RS with camera motion three times higher velocities than in the \emph{slow} case). The upward arrow means that a higher score is better. Our approach is significantly better than the considered methods, and with competitive results even for the GS case. Please see the text for details.}
	\label{tab:errors}
\end{table*}

\begin{table*}[t!]
\rotatebox[origin=bl]{0}{%
	\centering
	\resizebox{1\linewidth}{!}{%
		\begin{tabular}{lllllllllllllllllllllll}
			\toprule
			& \multicolumn{11}{c}{abs rel  $\downarrow$} & \multicolumn{11}{c}{$\delta < 1.25 \uparrow$}\\

 \cmidrule(lr){1-1} \cmidrule(lr){2-12} \cmidrule(lr){13-23}
rabbit & 0 & 1 & 2 & 3 & 4 & 5 & 6 & 7 & 8 & 9 & 10 & 0 & 1 & 2 & 3 & 4 & 5 & 6 & 7 & 8 & 9 & 10 \\
 \cmidrule(lr){1-1} \cmidrule(lr){2-12} \cmidrule(lr){13-23}
Jeon-CVPR \cite{jeon2015accurate} & 0.06 & 0.08 & 0.07 & 0.07 & 0.07 & 0.1 & 0.19 & 0.12 & 0.13 & 0.34 & 0.39 & \textbf{1.0} & \textbf{1.0} & \textbf{1.0} & \textbf{1.0} & \textbf{1.0} & \textbf{1.0} & 0.82 & \textbf{1.0} & 0.91 & 0.59 & 0.35 \\
OACC-Net \cite{OACC-Net} & 0.4 & 0.48 & 0.5 & 0.44 & 0.38 & 0.49 & 0.47 & 0.48 & 0.44 & 0.5 & 0.5 & 0.26 & 0.08 & 0.06 & 0.14 & 0.29 & 0.09 & 0.1 & 0.1 & 0.1 & 0.13 & 0.1 \\
Ours & \textbf{0.03} & \textbf{0.03} & \textbf{0.03} & \textbf{0.02} & \textbf{0.03} & \textbf{0.03} & \textbf{0.03} & \textbf{0.03} & \textbf{0.02} & \textbf{0.03} & \textbf{0.03} & \textbf{1.0} & \textbf{1.0} & \textbf{1.0} & \textbf{1.0} & \textbf{1.0} & \textbf{1.0} & \textbf{1.0} & \textbf{1.0} & \textbf{1.0} & \textbf{1.0} & \textbf{1.0} \\
 \cmidrule(lr){1-1} \cmidrule(lr){2-12} \cmidrule(lr){13-23}
table & 0 & 1 & 2 & 3 & 4 & 5 & 6 & 7 & 8 & 9 & 10 & 0 & 1 & 2 & 3 & 4 & 5 & 6 & 7 & 8 & 9 & 10 \\
 \cmidrule(lr){1-1} \cmidrule(lr){2-12} \cmidrule(lr){13-23}
Jeon-CVPR \cite{jeon2015accurate} & 0.03 & 0.03 & 0.04 & 0.03 & 0.05 & 0.03 & 0.05 & 0.09 & 0.05 & 0.17 & 0.07 & \textbf{1.0} & \textbf{1.0} & 0.99 & \textbf{1.0} & \textbf{1.0} & 1.0 & 0.97 & 0.96 & 0.99 & 0.76 & 0.94 \\
OACC-Net \cite{OACC-Net} & 0.17 & 0.21 & 0.2 & 0.19 & 0.19 & 0.2 & 0.19 & 0.24 & 0.15 & 0.25 & 0.2 & 0.69 & 0.6 & 0.64 & 0.64 & 0.63 & 0.59 & 0.67 & 0.55 & 0.79 & 0.5 & 0.65 \\
Ours & \textbf{0.02} & \textbf{0.02} & \textbf{0.02} & \textbf{0.03} & \textbf{0.02} & \textbf{0.03} & \textbf{0.04} & \textbf{0.02} & \textbf{0.04} & \textbf{0.03} & \textbf{0.04} & 0.995 & 0.995 & \textbf{0.99} & \textbf{1.0} & 0.99 & \textbf{1.0} & \textbf{1.0} & \textbf{0.99} & \textbf{0.99} & \textbf{0.98} & \textbf{1.0} \\
 \cmidrule(lr){1-1} \cmidrule(lr){2-12} \cmidrule(lr){13-23}
bedroom & 0 & 1 & 2 & 3 & 4 & 5 & 6 & 7 & 8 & 9 & 10 & 0 & 1 & 2 & 3 & 4 & 5 & 6 & 7 & 8 & 9 & 10 \\
 \cmidrule(lr){1-1} \cmidrule(lr){2-12} \cmidrule(lr){13-23}
Jeon-CVPR \cite{jeon2015accurate} & \textbf{0.02} & 0.03 & \textbf{0.02} & 0.04 & 0.06 & 0.03 & 0.07 & \textbf{0.02} & 0.11 & \textbf{0.03} & 0.07 & \textbf{1.0} & 1.0 & \textbf{1.0} & \textbf{1.0} & 0.97 & 1.0 & 0.99 & \textbf{1.0} & 0.89 & \textbf{1.0} & 0.94 \\
OACC-Net \cite{OACC-Net} & 0.03 & 0.05 & 0.03 & 0.06 & 0.1 & 0.05 & 0.13 & 0.03 & 0.13 & 0.05 & 0.13 & 1.0 & 0.98 & 1.0 & 0.99 & 0.93 & 0.98 & 0.8 & 1.0 & 0.77 & \textbf{1.0} & 0.79 \\
Ours & 0.03 & \textbf{0.03} & 0.03 & \textbf{0.03} & \textbf{0.03} & \textbf{0.03} & \textbf{0.04} & 0.03 & \textbf{0.04} & 0.04 & \textbf{0.05} & \textbf{1.0} & \textbf{1.0} & \textbf{1.0} & 0.999 & \textbf{1.0} & \textbf{1.0} & \textbf{1.0} & \textbf{1.0} & \textbf{1.0} & \textbf{1.0} & \textbf{0.99} \\
             \bottomrule
		\end{tabular}
	}
}
 \vspace*{-0.2cm}
\caption{Detailed reconstruction error metrics for three representative scenes ``rabbit", ``table" and ``bedroom" considering the eleven different motion scenarios (from 0 to 10) of the dataset. The upward arrow means that a higher score is better.}\vspace*{-0.5cm}
\label{tab:scenes}
\end{table*}

Despite the potential of rolling shutter plenoptic cameras, to the best of the authors' knowledge, all existing LF datasets are done assuming a global shutter hypothesis \cite{dansereau2019liff, pertuz2018focus, rerabek2016new, Vamsi2017}. Unfortunately, there is no public data available showing the rolling shutter effect on light-field images. Therefore, we have carefully designed and collected a new dataset with seven different synthetic scenes build on Blender, containing notably pseudo-real scenes created from Habitat-Matterport benchmark~\cite{ramakrishnan2021hm3d}. This new dataset (inspired by the HCI 4D LF benchmark \cite{honauer2017dataset}) is composed of four photo-realistic scenes from Matterport and three synthetic ones (as the examples depicted in \cref{fig:simu} and \cref{fig:qualit}). We provide, per scene, the following data: \vspace*{-0.2cm}
\begin{enumerate}[label=(\roman*)]
    \item Config files with camera settings and disparity ranges.\vspace*{-0.2cm}
    \item Different motion scenarios: \vspace*{-0.2cm}
    \begin{itemize}
        \item \textit{GS}: This is the static configuration. It allows to have a good measure of the performance difference with or without RS distortion by having the same scene in both scenarios. It is equivalent to a GS light field.
    	\item \textit{slow}: The motions affect the image enough to affect largely the perception of the scene geometry.
    	\item \textit{fast}: The linear and angular camera velocities are about three times more important than for the \textit{slow} motions. 
    \end{itemize} 
    We collect 11 light field sequences per scene (1 \textit{GS}, 5 \textit{slow}, 5 \textit{fast}). Please see the table in the supplementary with the velocity intervals for each motion scenario.\vspace{-0.2cm}
    \item Each light field is of dimension $9\times9\times512\times512\times3$, which is equivalent to a light field captured from a plenoptic camera with a $512\times512$ micro-lense array and $9\times9$ micro-images. \vspace{-0.2cm}
    \item A depth map corresponding to the geometry of the scene at middle time of exposition (\ie, the pose of the camera during the acquisition of the center line). \vspace{-0.2cm}
\end{enumerate}
We believe this dataset has the potential to help the evaluation and to promote futher investigation of RS applications for scene analysis with light fields. Visualizations and additional details of the dataset are given in the Supplementary material. 

\section{Experiments}

\paragraph{Metrics and competitors.}
We have selected two representative algorithms for comparison: the model-based approach of Jeon \etal \cite{jeon2015accurate}, and a recent learning-based 3D estimation from LF of Wang \etal \cite{OACC-Net}. The comparison is done in both GS and RS scenarios for all methods with the aim of fair conditions for the competitors.
Six commonly used metrics are selected for the evaluation \textit{abs rel}, \textit{abs diff}, \textit{RMS}, \textit{$\delta < 1.25$}, \textit{$\delta < 1.25^2$} and \textit{$\delta < 1.25^3$}. \textit{abs rel} is the absolute difference between the estimation and the ground truth (gt), normalized by the gt. \textit{abs diff} is the absolute difference between the estimation and the gt. \textit{RMS} is the Root Mean Square Error between the estimation and the gt. \textit{$\delta < 1.25$}, \textit{$\delta < 1.25^2$} and \textit{$\delta < 1.25^3$} are respectively the proportion of the points in a range of $1.25$ times the gt, $1.25^2$ times the gt and $1.25^3$ times the gt.

\subsection{Results}
\begin{figure}[t]
	\centering	
	\includegraphics[width=0.93\linewidth]{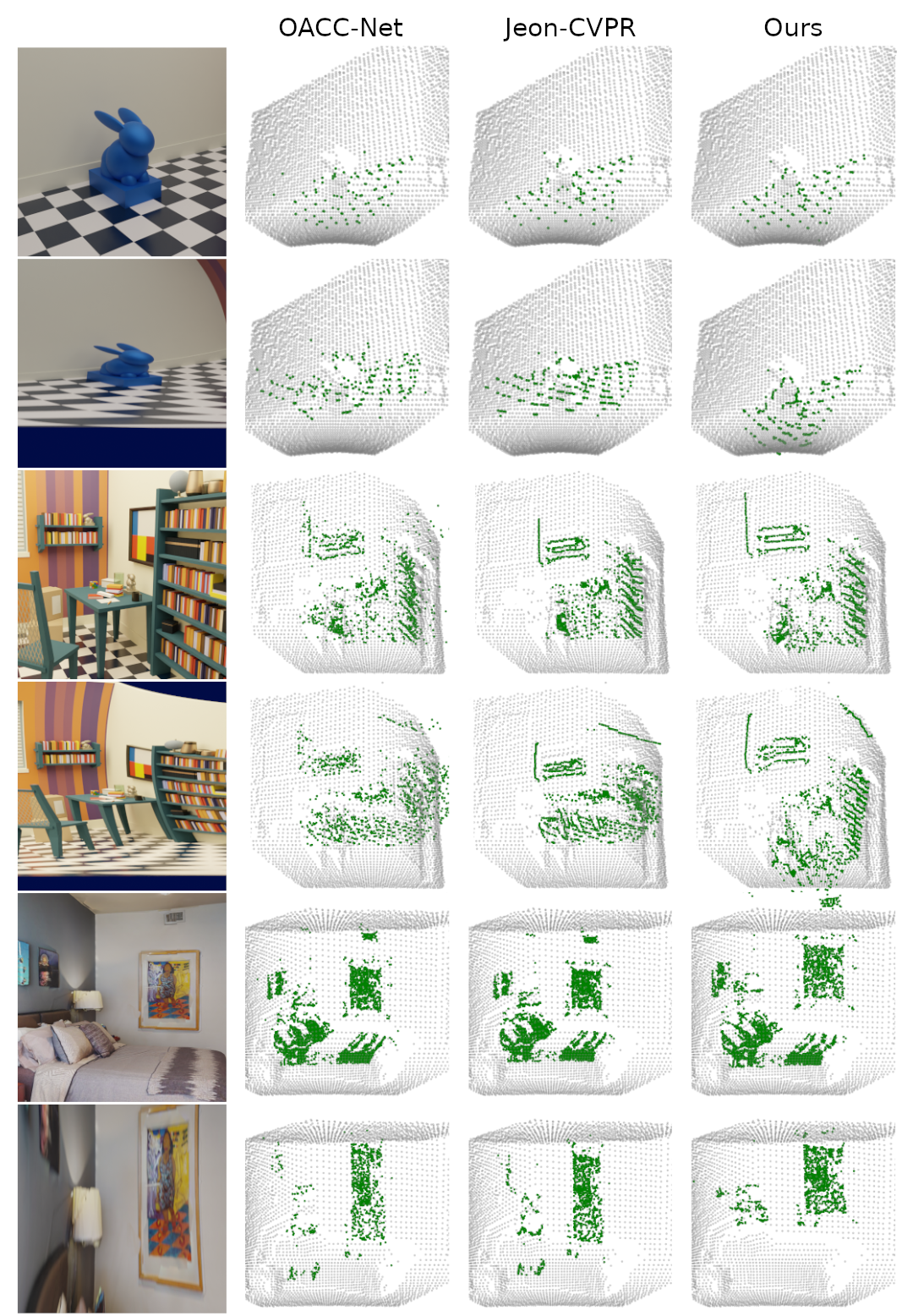}
	\caption{Some central views and associated point cloud reconstructions for the scenes and results shown in \cref{tab:scenes}. From right to left, OACC-Net \cite{OACC-Net}, Jeon-CVPR \cite{jeon2015accurate} and Ours. Ground truth points in gray and estimated in green.}\vspace*{-0.3cm}
	\label{fig:qualit}
\end{figure}

The evaluation and averaged metrics for all scenes (and different motion conditions) are shown in \cref{tab:errors}. We can observe the proposed method achieves the best scores overall in several of the considered metrics (e.g., ``abs rel" and ``abs diff"), and 
notably for all metrics of the ``fast'' sequences' split. We can also notice that it has even a competitive performance to the recent competitors in the GS scenario. This aspect will be further investigated in the ablation and sensitivity analysis. As we can observe, the two competitors perform far worse when motion is present.
The detailed metrics for three representative scenes considering the eleven light fields sequences per scene (1 \textit{GS}, 5 \textit{slow}, 5 \textit{fast}) are shown in \cref{tab:scenes}, where we can see that our method performs better in most cases. 
Please check some qualitative examples of the obtained shape reconstructions for these three scenes shown in \cref{fig:qualit}. We alternate, for these three scenes, the GS case and a RS case with high velocity (motion scenario number 9). We can clearly see the capacity of our algorithm to model the RS deformations.
In the scene ``bedroom", motion scenario 9, (the last line of \cref{fig:qualit}), one can clearly notice from visual inspection the compensation done on the painting (the rectangle is less stretched). Unfortunately, this qualitative observation is not highlighted in the detailed quantitative metrics \cref{tab:scenes}. Indeed, the painting is stretched in the estimation given by the competitors, but is still close to the wall plane, resulting in similar scores.
However, the proposed formulation is at least twice as accurate than the competitors for the other two scenes detailed in \cref{tab:scenes} for all motion profiles, accordingly to the average scores for all sequences shown in \cref{tab:errors}.
The detailed results for all sequences and scenes are included for completeness in the Supplementary materials due to page space limitations.

\begin{figure*}[t!]
	\centering
	\includegraphics[width=0.85\linewidth]{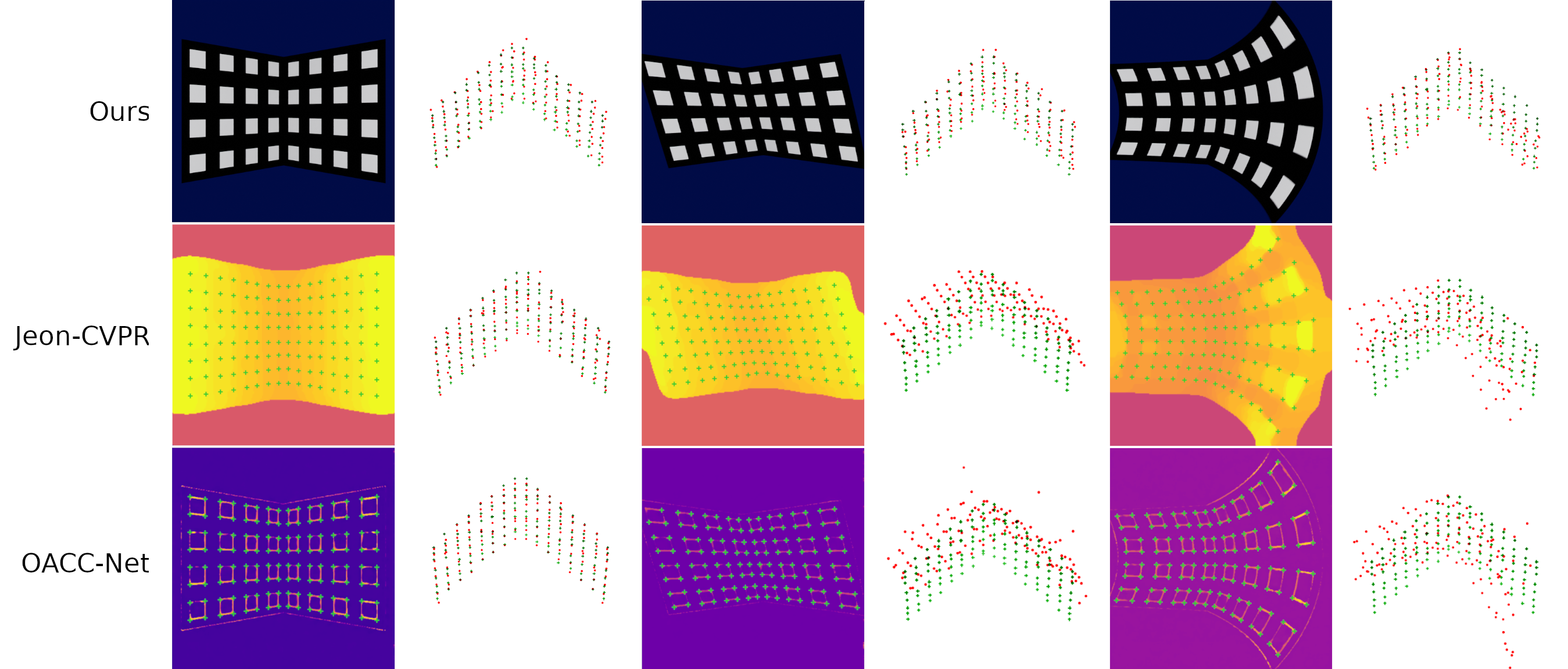}
	\caption{Qualitative examples of reconstruction for different motion scenarios for the ``chart" sequence. The "GS" scenario on the left. A "slow" scenario in the middle. A "fast" scenario on the right. - \textit{first column}: The central view of the LF, the disparity map of Jeon-CVPR \cite{jeon2015accurate}, the disparity map of OACC-Net \cite{OACC-Net}. - \textit{Second column}: The 3D point clouds (red dots) obtained for our method, Jeon-CVPR \cite{jeon2015accurate}, OACC-Net \cite{OACC-Net}. Despite the fact that the images look different, due to the rolling shutter effect, the reconstruction is supposed to give the same result (in green crosses in the point clouds).}
	\label{fig:compet}
\end{figure*}

Finally, we analyse the performance of the approaches in the easy to understand \textit``chart" scene as shown in the quantitative results from \cref{tab:errors_chart} and visualizations in \cref{fig:compet}. Similarly to all other scenes, it is composed of eleven light fields (1 \textit{GS}, 5 \textit{slow}, 5 \textit{fast}), where a double checkerboard pattern is joint in a $90^{\circ}$ angle configuration. Our method achieves the best scores for every metric in both the \textit{slow} and \textit{fast} scenarios. However, we can also obtain competitive results to both strong competitors in the case of GS. We can also notice that sometimes our obtained estimation is more accurate when the camera is moving slowly than when the camera is static. This will be discussed in the ablation study \cref{sec:abla}. \cref{tab:errors_chart} also indicates that our method slightly degrades with the augmentation of the camera speed, but it still considerably outperforms all the competitors in the \textit{fast} scenarios for the four first metrics. \cref{fig:compet} shows some qualitative examples of the three methods in the different scenarios and the associated point clouds. We can observe how our method is still capable of fitting the object shape even with the presence of RS and fast camera motions. Looking at the object 3D reconstruction results obtained by the other methods, we can clearly observe deformation effects caused by the misinterpretation of the RS checkerboard images. These degradation of performance can be explained if we observe that the computed disparity maps of the competitors map the distortions of the scene due to RS from the center view. They also interpret the movement of the camera between vertically distant views only as spatial disparity. Thus, if a point moves vertically downwards during acquisition, it will have a bigger disparity than it should (between two viewpoints, where the point is moving because of changes in point of view but also because of its own movement). Inversely, if a point moves vertically upwards during acquisition, it will have a smaller disparity than it should. These two effects contribute to degrade the performance of GS-designed algorithms in the estimation of the 3D geometry of the scene.

\begin{table*}[t!]
	\centering
	\resizebox{1.3\columnwidth}{!}{%
		\begin{tabular}{llllllllll}
			\toprule
			& \multicolumn{3}{c}{abs rel $\downarrow$} & \multicolumn{3}{c}{abs diff $\downarrow$} & \multicolumn{3}{c}{RMS $\downarrow$}\\
			\cmidrule(lr){2-4} \cmidrule(lr){5-7} \cmidrule(lr){8-10}
            Method & GS & slow & fast & GS & slow & fast& GS & slow & fast\\
        	\cmidrule(lr){1-1} \cmidrule(lr){2-4} \cmidrule(lr){5-7} \cmidrule(lr){8-10}
            Jeon-CVPR \cite{jeon2015accurate} &       \textbf{0.003} &      0.013 &      0.049 &      \textbf{8.464} &     30.293 &     76.824 &  \textbf{17.489} &     47.647 &    129.720\\
            OACC-Net \cite{OACC-Net} &      \textbf{0.003} &      0.013 &      0.051 &     12.214 &     30.882 &     79.938 &     26.197 &     54.799 &    140.215\\
            Ours &      0.004 &      \textbf{0.003} &      \textbf{0.003} &     13.692 &     \textbf{15.395} &     \textbf{23.754} &     22.146 &     \textbf{25.327} &     \textbf{44.791}\\
            \bottomrule
            \toprule
            & \multicolumn{3}{c}{$\delta < 1.25 \uparrow$} & \multicolumn{3}{c}{$\delta < 1.25^2 \uparrow$} & \multicolumn{3}{c}{$\delta < 1.25^3 \uparrow$}\\
            \cmidrule(lr){2-4} \cmidrule(lr){5-7} \cmidrule(lr){8-10}
            Method & GS & slow & fast& GS & slow & fast& GS & slow & fast\\
            \cmidrule(lr){1-1} \cmidrule(lr){2-4} \cmidrule(lr){5-7} \cmidrule(lr){8-10}
             Jeon-CVPR \cite{jeon2015accurate} &      \textbf{1.000 }&      0.923 &      0.745 &      \textbf{1.000} &      0.992 &      0.898 &      \textbf{1.000} &      0.995 &      0.991 \\
             OACC-Net \cite{OACC-Net}  &      \textbf{1.000} &      0.922 &      0.730 &      \textbf{1.000} &      0.993 &      0.895 &      \textbf{1.000} &      0.996 &      0.939\\
             Ours &      0.988 &      \textbf{0.982} &      \textbf{0.973} &      0.996 &      \textbf{0.999} &      \textbf{0.995} &      \textbf{1.000} &      \textbf{1.000} &      \textbf{0.998}\\
             \bottomrule
		\end{tabular}
  }
 \vspace{0.1cm}
 \caption{Detailed reconstruction error metrics in different scenarios for the ``chart" sequence: \textit{GS} (global shutter, equivalent to a static camera scenario), \textit{slow} (RS with small camera linear and angular velocities), and \textit{fast} (RS with camera motion three times higher velocities than in the \emph{slow} case). The upward arrow means that a higher score is better. Our approach performed significantly better than the two recent considered methods.}
	\label{tab:errors_chart}
\end{table*}

\subsection{Ablation study}\label{sec:abla}

We performed different ablation studies in order to evaluate the relevance of the different parts of the method. In the first ablation, we retained two major components for evaluation, the contribution of i) linear initialization strategy (No Init.), and ii) the regularization (No Reg.) as shown in \cref{tab:ablation}. For the ablation of the initialization, we initialized the optimization \cref{eq:error_func2} with all the points clustered in a position near the center of mass of the point cloud we should have found with the linear initialization. We show in \cref{tab:ablation} that, even after convergence, the solution is still far from correct. For the ablation of the regularization, we see that the method without the regularization gives worst results in the RS scenarios. These evaluations confirm the importance of these components in the designed method.

A second ablation study was designed to evaluate the performance of our method without the RS modelling (Ours No RS) depicted in \cref{tab:ablation_1}.
By modeling the RS effect we also have additional degrees of freedom that lead to a slight degradation of the results when compared to a GS scheme for the GS scenes. We performed an evaluation to verify the effect of constraining the dynamic degrees of freedom ($\Omega=0$ and $\mathbf{v}=\mathbf{0}$) in case of GS would result in the estimation. The results in \cref{tab:ablation_1} show an improvement on all the metrics of up to about 6\%. This concurs with the aforementioned hypothesis. The obtained performance is on par with the competitors which are specifically designed for GS settings.

\subsection{Discussion}
From the experiments, we can observe that our method is capable of handling different camera motions and provides improved scene structure estimates. The proposed model is designed to handle rigid scenes, yet it can estimate the structure and motion parameters for 3D scene points independently if at least four image points are available, \ie, to compute a “3D scene flow” from a single LF image. We assumed rigidity in order to compute a common set of dynamic parameters to each point (corresponding to a camera motion in a rigid scene). We believe our strategy could be also extended to handle scenes with dynamic objects independently (or non-rigid) with multiple camera motion hypotheses. The adopted RS projection also assumes that both linear and angular velocities to be uniform during the LF image acquisition (i.e., zero acceleration). However, RS devices, while having a sequential acquisition, usually have a small time of total exposure per frame (about 0.1 s for a 4K image). Therefore the assumption of constant camera speeds during the frame acquisition holds in typical motion-scene scale scenarios. Nevertheless, the proposed approach could still be applied for accelerated motions with a piece-wise decomposition of the plenoptic image in horizontal bands. Such a strategy of piece-wise decomposition in horizontal bands for classic images has been investigated in \cite{magerand2010generic} for a classic monocular RS sensor. The motion and shape estimation could then be done at different time instants and allow to recover more complex scenes (\eg, non-rigid) and motion scenarios. 

 \begin{table} [!t]
    \center
    \resizebox{0.5\columnwidth}{!}{%
	\begin{tabular}{lllllll}
        \toprule
		& \multicolumn{3}{c}{RMS $\downarrow$}\\
		\cmidrule(lr){2-4}
        Abl. & GS & slow & fast\\
        \cmidrule(lr){1-1} \cmidrule(lr){2-4}
        No Init. &    0.243 &    0.242 &  0.240 \\
        No Reg. &    $\textbf{0.045}$ &     0.060 &     0.086\\
        Full &     0.046 &     $\textbf{0.051}$ &     $\textbf{0.064}$\\
        \bottomrule
        \toprule
        & \multicolumn{3}{c}{$\delta_1 < 1.25 \uparrow$}\\
        \cmidrule(lr){2-4}
        Abl. & GS & slow & fast\\
        \cmidrule(lr){1-1} \cmidrule(lr){2-4}
        No Init. &      0.650 &      0.646 &      0.630 \\
        No Reg. &      $\textbf{0.969}$ &      0.950 &      0.895  \\
        Full &      0.958 &      $\textbf{0.961}$ &     $\textbf{0.949}$ \\
        \bottomrule
	\end{tabular}    }
    
    \caption{Reconstruction errors for the ablation study of our method for the initialization and regularization steps.}
    \vspace*{0.3cm}
    \label{tab:ablation}

	\resizebox{\columnwidth}{!}{%
		\begin{tabular}{lllll}
			\toprule
			Abl. & abs rel $\downarrow$ & abs diff $\downarrow$ & RMS $\downarrow$ & $\delta < 1.25 \uparrow $ \\
			\cmidrule(lr){1-1} \cmidrule(lr){2-5}
            Jeon-CVPR \cite{jeon2015accurate} & 0.040 & 0.027 & 0.035 & 0.993 \\
            Ours Full & 0.040 & 0.031 & 0.046 & 0.958 \\
            Ours No RS & 0.040 & 0.029 & 0.041 & 0.976 \\
            \bottomrule
		\end{tabular}	}
 \caption{Ablation study of the dynamic motion parameters with a static GS scene.}\vspace*{-0.3cm}
	\label{tab:ablation_1}
\end{table}

\section{Conclusion}
In this paper, we proposed a projection model for a light-field camera equipped with a rolling shutter sensor. This model allows us to jointly estimate the shape and motion on unknown scenes from a single light field image. The approach has been evaluated on different motions and 3D scenes. Furthermore, it does not suffer from shape/motion ambiguity thanks to the relatively reasonable assumption of a row-wise GS. To fill the lack of publicly available rolling-shutter LF data, we created a dataset that includes simulated photo-realistic light fields with different motion scenarios, and we will make it publicly available. We plan to build upon this model to generate denser depth maps and extend the motion estimations to non-rigid scenes. Our proposed model shows improved 3D scene geometry estimates, and we believe that it will inspire further research in this area, notably for applications in the context of robot vision, manipulation and micro-robotics.

\paragraph{Acknowledgements.} The authors would like to thank the funding from the French ``Investissements d’Avenir'' program, project ISITE-BFC, contract ANR-15-IDEX-03, by the Conseil Régional BFC from the project ANER-MOVIS and by "Grand Prix Scientifique 2018, Fond. Ch. Defforey-Institut de France".

\cleardoublepage
\balance

%%%%%%%%% REFERENCES
{\small
\bibliography{ma.bbl}

\begin{thebibliography}{10}\itemsep=-1pt

\bibitem{adelson1992single}
Edward~H Adelson and John~YA Wang.
\newblock Single lens stereo with a plenoptic camera.
\newblock {\em IEEE Trans. Pattern Anal. Mach. Intell.}, 14(2):99--106, 1992.

\bibitem{Vamsi2017}
Vamsi~Kiran Adhikarla, Marek Vinkler, Denis Sumin, Rafał Mantiuk, Karol
  Myszkowski, Hans-Peter Seidel, and Piotr Didyk.
\newblock Towards a quality metric for dense light fields.
\newblock In {\em IEEE Conf. Comput. Vis. Pattern Recog.}, 2017.

\bibitem{ait2006simultaneous}
Omar Ait-Aider, Nicolas Andreff, Jean~Marc Lavest, and Philippe Martinet.
\newblock Simultaneous object pose and velocity computation using a single view
  from a rolling shutter camera.
\newblock In {\em Eur. Conf. Comput. Vis.}, 2006.

\bibitem{ait2009structure}
Omar Ait-Aider and Fran{\c{c}}ois Berry.
\newblock Structure and kinematics triangulation with a rolling shutter stereo
  rig.
\newblock In {\em Int. Conf. Comput. Vis.}, 2009.

\bibitem{bok2016geometric}
Yunsu Bok, Hae-Gon Jeon, and In~So Kweon.
\newblock Geometric calibration of micro-lens-based light field cameras using
  line features.
\newblock {\em IEEE Trans. Pattern Anal. Mach. Intell.}, 39(2):287--300, 2016.

\bibitem{bolles1987epipolar}
Robert~C Bolles, H~Harlyn Baker, and David~H Marimont.
\newblock Epipolar-plane image analysis: An approach to determining structure
  from motion.
\newblock {\em Int. J. Comput. Vis.}, 1(1):7--55, 1987.

\bibitem{conti2020dense}
Caroline Conti, Lu{\'\i}s~Ducla Soares, and Paulo Nunes.
\newblock Dense light field coding: A survey.
\newblock {\em Access}, 8:49244--49284, 2020.

\bibitem{dansereau2004gradient}
Don Dansereau and Len Bruton.
\newblock Gradient-based depth estimation from 4d light fields.
\newblock In {\em Int. Symposium on Circuits and Systems}. IEEE, 2004.

\bibitem{dansereau2019liff}
Donald~G. Dansereau, Bernd Girod, and Gordon Wetzstein.
\newblock {LiFF}: Light field features in scale and depth.
\newblock In {\em IEEE Conf. Comput. Vis. Pattern Recog.}, 2019.

\bibitem{fan2021sunet}
Bin Fan, Yuchao Dai, and Mingyi He.
\newblock Sunet: symmetric undistortion network for rolling shutter correction.
\newblock In {\em Int. Conf. Comput. Vis.}, 2021.

\bibitem{ferreira2016fast}
Rodrigo Ferreira and Nuno Goncalves.
\newblock Fast and accurate micro lenses depth maps for multi-focus light field
  cameras.
\newblock In {\em German Conf. on Pattern Recog.} Springer, 2016.

\bibitem{georgiev2010focused}
Todor~G Georgiev and Andrew Lumsdaine.
\newblock Focused plenoptic camera and rendering.
\newblock {\em J. of Electronic Imaging}, 19(2):021106, 2010.

\bibitem{heber2016convolutional}
Stefan Heber and Thomas Pock.
\newblock Convolutional networks for shape from light field.
\newblock In {\em IEEE Conf. Comput. Vis. Pattern Recog.}, 2016.

\bibitem{honauer2017dataset}
Katrin Honauer, Ole Johannsen, Daniel Kondermann, and Bastian Goldluecke.
\newblock A dataset and evaluation methodology for depth estimation on 4d light
  fields.
\newblock In {\em Asian Conf. on Comput. Vis.}, pages 19--34. Springer, 2017.

\bibitem{ihrke2016principles}
Ivo Ihrke, John Restrepo, and Lois Mignard-Debise.
\newblock Principles of light field imaging: Briefly revisiting 25 years of
  research.
\newblock {\em Sign. Proc. Magazine}, 33(5):59--69, 2016.

\bibitem{jeon2015accurate}
Hae-Gon Jeon, Jaesik Park, Gyeongmin Choe, Jinsun Park, Yunsu Bok, Yu-Wing Tai,
  and In So~Kweon.
\newblock Accurate depth map estimation from a lenslet light field camera.
\newblock In {\em IEEE Conf. Comput. Vis. Pattern Recog.}, 2015.

\bibitem{karpenko2011digital}
Alexandre Karpenko, David Jacobs, Jongmin Baek, and Marc Levoy.
\newblock Digital video stabilization and rolling shutter correction using
  gyroscopes.
\newblock {\em CSTR}, 1(2):13, 2011.

\bibitem{LAO20181}
Yizhen Lao, Omar Ait-Aider, and Helder Araujo.
\newblock Robustified structure from motion with rolling-shutter camera using
  straightness constraint.
\newblock {\em Pattern Recognition Letters}, 111:1--8, 2018.

\bibitem{lao2021solving}
Yizhen Lao, Omar Ait-Aider, and Adrien Bartoli.
\newblock Solving rolling shutter 3d vision problems using analogies with
  non-rigidity.
\newblock {\em Int. J. Comput. Vis.}, 129(1):100--122, 2021.

\bibitem{liang2008analysis}
Chia-Kai Liang, Li-Wen Chang, and Homer~H Chen.
\newblock Analysis and compensation of rolling shutter effect.
\newblock {\em IEEE Trans. Image Process.}, 17(8):1323--1330, 2008.

\bibitem{lin2015depth}
Haiting Lin, Can Chen, Sing~Bing Kang, and Jingyi Yu.
\newblock Depth recovery from light field using focal stack symmetry.
\newblock In {\em Int. Conf. Comput. Vis.}, 2015.

\bibitem{magerand2010generic}
Ludovic Magerand and Adrien Bartoli.
\newblock A generic rolling shutter camera model and its application to dynamic
  pose estimation.
\newblock In {\em Int. symposium on 3D Data Proc., Visualiz. and Transmis.},
  2010.

\bibitem{meilland2013unified}
Maxime Meilland, Tom Drummond, and Andrew~I Comport.
\newblock A unified rolling shutter and motion blur model for 3d visual
  registration.
\newblock In {\em Int. Conf. Comput. Vis.}, 2013.

\bibitem{meingast2005geometric}
Marci Meingast, Christopher Geyer, and Shankar Sastry.
\newblock Geometric models of rolling-shutter cameras.
\newblock {\em arXiv preprint cs/0503076}, 2005.

\bibitem{ng2005light}
Ren Ng, Marc Levoy, Mathieu Br{\'e}dif, Gene Duval, Mark Horowitz, and Pat
  Hanrahan.
\newblock {\em Light field photography with a hand-held plenoptic camera}.
\newblock PhD thesis, Stanford University, 2005.

\bibitem{pertuz2018focus}
Said Pertuz, Edith Pulido-Herrera, and Joni-Kristian Kamarainen.
\newblock Focus model for metric depth estimation in standard plenoptic
  cameras.
\newblock {\em J. of Photogrammetry and Remote Sensing}, 144:38--47, 2018.

\bibitem{perwass2012single}
Christian Perwass and Lennart Wietzke.
\newblock Single lens 3d-camera with extended depth-of-field.
\newblock In {\em Human Vis. and Elect. imaging}. SPIE, 2012.

\bibitem{ramakrishnan2021hm3d}
Santhosh~Kumar Ramakrishnan, Aaron Gokaslan, Erik Wijmans, Oleksandr Maksymets,
  Alexander Clegg, John~M Turner, Eric Undersander, Wojciech Galuba, Andrew
  Westbury, Angel~X Chang, Manolis Savva, Yili Zhao, and Dhruv Batra.
\newblock Habitat-matterport 3d dataset ({HM}3d): 1000 large-scale 3d
  environments for embodied ai.
\newblock In {\em Adv. Neural Inform. Process. Syst.}, 2021.

\bibitem{rerabek2016new}
Martin Rerabek and Touradj Ebrahimi.
\newblock New light field image dataset.
\newblock In {\em Int. Confe. on Qual. of Multimed. Exp.}, 2016.

\bibitem{saurer2013rolling}
Olivier Saurer, Kevin Koser, Jean-Yves Bouguet, and Marc Pollefeys.
\newblock Rolling shutter stereo.
\newblock In {\em Int. Conf. Comput. Vis.}, 2013.

\bibitem{shin2018epinet}
Changha Shin, Hae-Gon Jeon, Youngjin Yoon, In~So Kweon, and Seon~Joo Kim.
\newblock Epinet: A fully-convolutional neural network using epipolar geometry
  for depth from light field images.
\newblock In {\em IEEE Conf. Comput. Vis. Pattern Recog.}, 2018.

\bibitem{NEURIPS2021_a11ce019}
Vincent Sitzmann, Semon Rezchikov, Bill Freeman, Josh Tenenbaum, and Fredo
  Durand.
\newblock Light field networks: Neural scene representations with
  single-evaluation rendering.
\newblock In {\em Adv. Neural Inform. Process. Syst.}, 2021.

\bibitem{srinivasan2017light}
Pratul~P Srinivasan, Ren Ng, and Ravi Ramamoorthi.
\newblock Light field blind motion deblurring.
\newblock In {\em IEEE Conf. Comput. Vis. Pattern Recog.}, 2017.

\bibitem{tao2015depth}
Michael~W Tao, Pratul~P Srinivasan, Jitendra Malik, Szymon Rusinkiewicz, and
  Ravi Ramamoorthi.
\newblock Depth from shading, defocus, and correspondence using light-field
  angular coherence.
\newblock In {\em IEEE Conf. Comput. Vis. Pattern Recog.}, 2015.

\bibitem{OACC-Net}
Yingqian Wang, Longguang Wang, Zhengyu Liang, Jungang Yang, Wei An, and Yulan
  Guo.
\newblock Occlusion-aware cost constructor for light field depth estimation.
\newblock In {\em IEEE Conf. Comput. Vis. Pattern Recog.}, 2022.

\bibitem{wanner2012globally}
Sven Wanner and Bastian Goldluecke.
\newblock Globally consistent depth labeling of 4d light fields.
\newblock In {\em IEEE Conf. Comput. Vis. Pattern Recog.}, 2012.

\bibitem{wu2017light}
Gaochang Wu, Belen Masia, Adrian Jarabo, Yuchen Zhang, Liangyong Wang, Qionghai
  Dai, Tianyou Chai, and Yebin Liu.
\newblock Light field image processing: An overview.
\newblock {\em J. of Selected Topics in Sign. Proc.}, 11(7):926--954, 2017.

\bibitem{zeller2016depth}
Niclas Zeller, Franz Quint, and Uwe Stilla.
\newblock Depth estimation and camera calibration of a focused plenoptic camera
  for visual odometry.
\newblock {\em J. of Photogrammetry and Remote Sensing}, 118:83--100, 2016.

\bibitem{zhang2018generic}
Qi Zhang, Chunping Zhang, Jinbo Ling, Qing Wang, and Jingyi Yu.
\newblock A generic multi-projection-center model and calibration method for
  light field cameras.
\newblock {\em IEEE Trans. Pattern Anal. Mach. Intell.}, 41(11):2539--2552,
  2018.

\bibitem{zhang2016robust}
Shuo Zhang, Hao Sheng, Chao Li, Jun Zhang, and Zhang Xiong.
\newblock Robust depth estimation for light field via spinning parallelogram
  operator.
\newblock {\em Comput. Vis. and Image Underst.}, 145:148--159, 2016.

\end{thebibliography}
}

%---Supplementary--------------------------------------------

\onecolumn
\begin{center}
    \section*{[Supplementary material WACV 2024]:\\
Joint 3D Shape and Motion Estimation from Rolling Shutter Light-Field Images}
\end{center}

\setcounter{section}{0} 

In this supplementary material to our paper, we provide additional visualizations of the sequences from our dataset with rolling shutter light-field images, as well as more details about the projection model presented in the paper. 

\section{Rolling Shutter Light-Field Dataset}
\begin{figure}[ht!]
        \centering
	    \includegraphics[width=\linewidth]{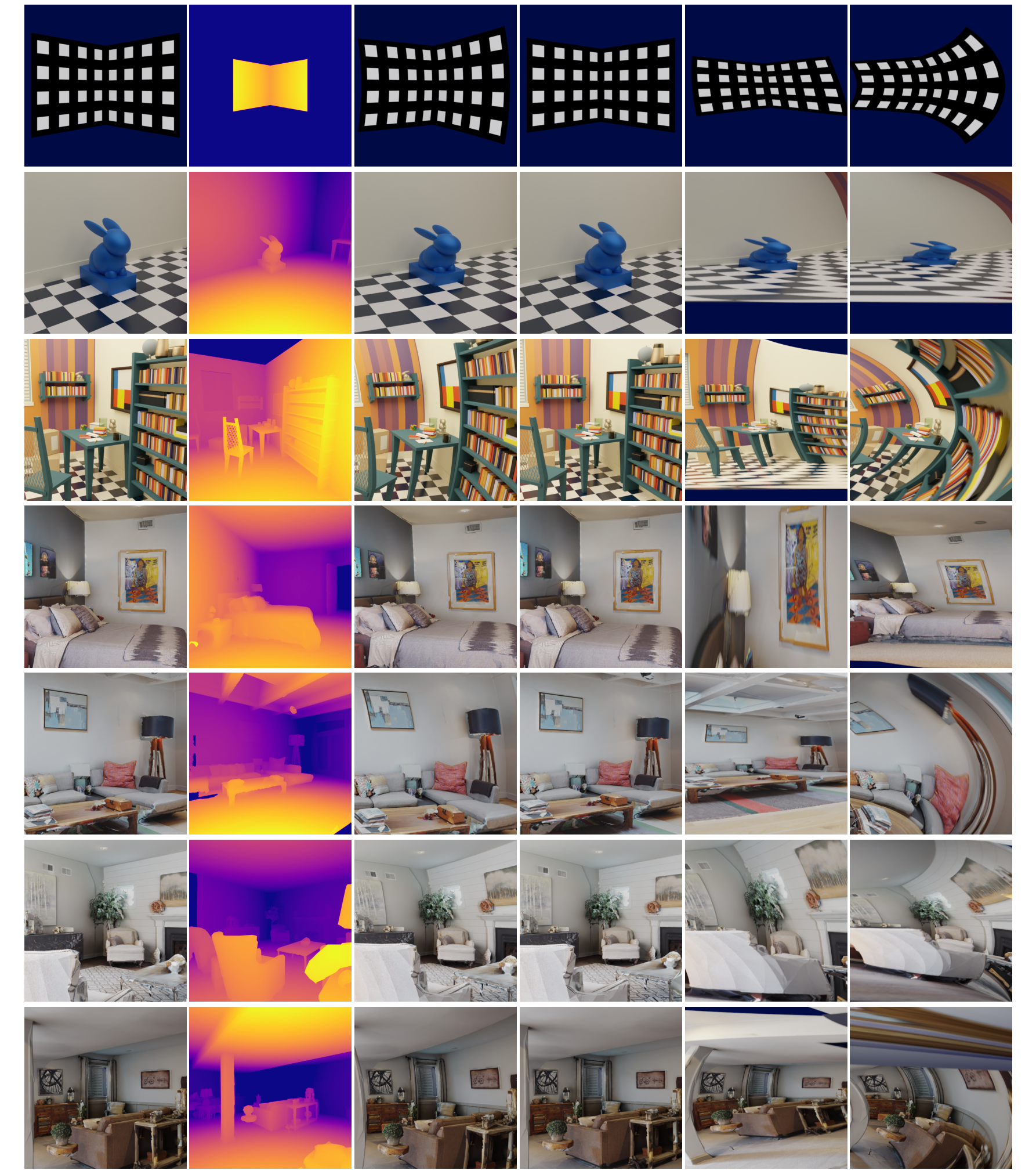}
	    \caption{This figure is already presented in the main paper and is duplicated here with bigger dimensions - Visualizations of the center views from each of the seven different scenes. The four bottom rows are generated from samples of the Habitat-Matterport benchmark ~\cite{ramakrishnan2021hm3d}. The first column shows the GS scenario; the second column shows the associated depth maps, and the subsequent columns are from different motion scenarios (numbers 1, 2, 9 and 10 in \cref{tab:velo}).}
	    \label{fig:dataset}
\end{figure}
Visualizations of some sequences scenes from our dataset (discussed in Section 4) are shown in \cref{fig:dataset}. For each scene, we provide eleven velocity scenarios from which every middle-exposition time position is similar, \ie the center line of pixel is the same for every image of the same scene. All the scenes share the same eleven velocity profiles, but still possess different deformations due to the RS effect since the center of rotation of the camera is always different. Please notice the far right column in \cref{fig:dataset}
where the same movement creates some ``squishing'' in some sequences (\eg second and fourth rows) and some ``stretching'' in others (\eg third and fifth rows).
The velocities for each scenarios are presented in \cref{tab:velo}.

\begin{table}[h]
	\centering
	\resizebox{0.8\columnwidth}{!}{%
		\begin{tabular}{lccccc}
			\toprule
            Scenario number & 1 & 2 & 3 & 4 &  5 \\
			\cmidrule(lr){2-6}
            Rotations (euler angles) & [0,0,$\pi$/12] & [0,0,0] & [-$\pi$/18,0,0] & [$\pi$/18,$\pi$/18,0] & [0,0,$\pi$/12] \\
            Translations & [0,0,0] & [0,-0.2,0] & [0,-0.05,0.05] & [0,0,0.2] &        [0,-0.2,0] \\
            \bottomrule
            \toprule
            Scenario number & 6 & 7 & 8 & 9 & 10 \\
			\cmidrule(lr){2-6}
            Rotations (euler angles) & [0,0,$\pi$/3] & [0,0,0] & [-$\pi$/3,0,0] & [2$\pi$/9,0,0] & [0,0,$\pi$/2] \\
            Translations & [0,0,0] & [0,-0.8,0] & [0,0.4,0.2] & [0.4,-1.6,-0.8] & [0,-0.8,0] \\
             \bottomrule
		\end{tabular}
	}
 \vspace{0.1cm}
 \caption{Different velocity scenarios, given in radians per frames and meters per frames. The scenarios from 1 to 5 are the \textit{slow} scenarios and the ones from 6 to 10 are the \textit{fast} scenarios.}
	\label{tab:velo}
\end{table}

All sequences were created using Blender and the render engine Cycles. The cameras have a $50mm$ focal length, and they are placed in a plane normal to the $z$ axis organized in a $9 \times 9$ grid with  $6mm$ between each  view point. They are all oriented in a way that they have their optical axes passing through the point $[0,0,7m]$. We also provide $1024x1024$ depth maps generated from the camera placed at the same position that of the center view, but with a $25mm$ focal length. Since the scene is moving with respect to the camera, it is necessary to have a wider field of view as more of the scene can be seen in the RS scenarios. The depth maps are normalized for a range from $0m$ to $7m$. Some vizualisations of the depth maps are shown in the second column of \cref{fig:dataset}. The detailed results mentioned in the main paper for all sequences and scenes are presented \cref{tab:scenes}.

\begin{table}[t!]
\centering
\resizebox{1\linewidth}{!}{
\begin{tabular}{lllllllllllllllllllllll}
\toprule
	& \multicolumn{11}{c}{abs rel $\downarrow$} & \multicolumn{11}{c}{$\delta < 1.25 \uparrow$}\\
    \cmidrule(lr){1-1} \cmidrule(lr){2-12} \cmidrule(lr){13-23}
    rabbit & 0 & 1 & 2 & 3 & 4 & 5 & 6 & 7 & 8 & 9 & 10 & 0 & 1 & 2 & 3 & 4 & 5 & 6 & 7 & 8 & 9 & 10 \\
	\cmidrule(lr){1-1} \cmidrule(lr){2-12} \cmidrule(lr){13-23}
    Jeon-CVPR \cite{jeon2015accurate} & 0.06 & 0.08 & 0.07 & 0.07 & 0.07 & 0.1 & 0.19 & 0.12 & 0.13 & 0.34 & 0.39 & $\textbf{1.0}$ & $\textbf{1.0}$ & $\textbf{1.0}$ & $\textbf{1.0}$ & $\textbf{1.0}$ & $\textbf{1.0}$ & 0.82 & $\textbf{1.0}$ & 0.91 & 0.59 & 0.35 \\
    OACC-Net \cite{OACC-Net} & 0.4 & 0.48 & 0.5 & 0.44 & 0.38 & 0.49 & 0.47 & 0.48 & 0.44 & 0.5 & 0.5 & 0.26 & 0.08 & 0.06 & 0.14 & 0.29 & 0.09 & 0.1 & 0.1 & 0.1 & 0.13 & 0.1 \\
    Ours & $\textbf{0.03}$ & $\textbf{0.03}$ & $\textbf{0.03}$ & $\textbf{0.02}$ & $\textbf{0.03}$ & $\textbf{0.03}$ & $\textbf{0.03}$ & $\textbf{0.03}$ & $\textbf{0.02}$ & $\textbf{0.03}$ & $\textbf{0.03}$ & $\textbf{1.0}$ & $\textbf{1.0}$ & $\textbf{1.0}$ & $\textbf{1.0}$ & $\textbf{1.0}$ & $\textbf{1.0}$ & $\textbf{1.0}$ & $\textbf{1.0}$ & $\textbf{1.0}$ & $\textbf{1.0}$ & $\textbf{1.0}$ \\
    \cmidrule(lr){1-1} \cmidrule(lr){2-12} \cmidrule(lr){13-23}
    table & 0 & 1 & 2 & 3 & 4 & 5 & 6 & 7 & 8 & 9 & 10 & 0 & 1 & 2 & 3 & 4 & 5 & 6 & 7 & 8 & 9 & 10 \\
    \cmidrule(lr){1-1} \cmidrule(lr){2-12} \cmidrule(lr){13-23}
    Jeon-CVPR \cite{jeon2015accurate} & 0.03 & 0.03 & 0.04 & 0.03 & 0.05 & 0.03 & 0.05 & 0.09 & 0.05 & 0.17 & 0.07 & $\textbf{1.0}$ & $\textbf{1.0}$ & 0.99 & $\textbf{1.0}$ & $\textbf{1.0}$ & 1.0 & 0.97 & 0.96 & 0.99 & 0.76 & 0.94 \\
    OACC-Net \cite{OACC-Net} & 0.17 & 0.21 & 0.2 & 0.19 & 0.19 & 0.2 & 0.19 & 0.24 & 0.15 & 0.25 & 0.2 & 0.69 & 0.6 & 0.64 & 0.64 & 0.63 & 0.59 & 0.67 & 0.55 & 0.79 & 0.5 & 0.65 \\
    Ours & $\textbf{0.02}$ & $\textbf{0.02}$ & $\textbf{0.02}$ & $\textbf{0.03}$ & $\textbf{0.02}$ & $\textbf{0.03}$ & $\textbf{0.04}$ & $\textbf{0.02}$ & $\textbf{0.04}$ & $\textbf{0.03}$ & $\textbf{0.04}$ & 0.995 & 0.995 & $\textbf{0.99}$ & $\textbf{1.0}$ & 0.99 & $\textbf{1.0}$ & $\textbf{1.0}$ & $\textbf{0.99}$ & $\textbf{0.99}$ & $\textbf{0.98}$ & $\textbf{1.0}$ \\
    \cmidrule(lr){1-1} \cmidrule(lr){2-12} \cmidrule(lr){13-23}
    bedroom & 0 & 1 & 2 & 3 & 4 & 5 & 6 & 7 & 8 & 9 & 10 & 0 & 1 & 2 & 3 & 4 & 5 & 6 & 7 & 8 & 9 & 10 \\
    \cmidrule(lr){1-1} \cmidrule(lr){2-12} \cmidrule(lr){13-23}
    Jeon-CVPR \cite{jeon2015accurate} & $\textbf{0.02}$ & 0.03 & $\textbf{0.02}$ & 0.04 & 0.06 & 0.03 & 0.07 & $\textbf{0.02}$ & 0.11 & $\textbf{0.03}$ & 0.07 & $\textbf{1.0}$ & 1.0 & $\textbf{1.0}$ & $\textbf{1.0}$ & 0.97 & 1.0 & 0.99 & $\textbf{1.0}$ & 0.89 & $\textbf{1.0}$ & 0.94 \\
    OACC-Net \cite{OACC-Net} & 0.03 & 0.05 & 0.03 & 0.06 & 0.1 & 0.05 & 0.13 & 0.03 & 0.13 & 0.05 & 0.13 & 1.0 & 0.98 & 1.0 & 0.99 & 0.93 & 0.98 & 0.8 & 1.0 & 0.77 & $\textbf{1.0}$ & 0.79 \\
    Ours & 0.03 & $\textbf{0.03}$ & 0.03 & $\textbf{0.03}$ & $\textbf{0.03}$ & $\textbf{0.03}$ & $\textbf{0.04}$ & 0.03 & $\textbf{0.04}$ & 0.04 & $\textbf{0.05}$ & $\textbf{1.0}$ & $\textbf{1.0}$ & $\textbf{1.0}$ & 0.999 & $\textbf{1.0}$ & $\textbf{1.0}$ & $\textbf{1.0}$ & $\textbf{1.0}$ & $\textbf{1.0}$ & $\textbf{1.0}$ & $\textbf{0.99}$ \\
    \cmidrule(lr){1-1} \cmidrule(lr){2-12} \cmidrule(lr){13-23}
    couch & 0 & 1 & 2 & 3 & 4 & 5 & 6 & 7 & 8 & 9 & 10 & 0 & 1 & 2 & 3 & 4 & 5 & 6 & 7 & 8 & 9 & 10 \\
    \cmidrule(lr){1-1} \cmidrule(lr){2-12} \cmidrule(lr){13-23}
    Jeon-CVPR \cite{jeon2015accurate} & $\textbf{0.02}$ & $\textbf{0.03}$ & $\textbf{0.03}$ & $\textbf{0.03}$ & $\textbf{0.03}$ & $\textbf{0.02}$ & $\textbf{0.06}$ & 0.05 & $\textbf{0.06}$ & 0.05 & $\textbf{0.06}$ & $\textbf{1.0}$ & $\textbf{1.0}$ & $\textbf{1.0}$ & $\textbf{1.0}$ & $\textbf{1.0}$ & $\textbf{1.0}$ & $\textbf{1.0}$ & $\textbf{1.0}$ & $\textbf{1.0}$ & $\textbf{0.98}$ & $\textbf{1.0}$ \\
    OACC-Net \cite{OACC-Net} & 0.04 & 0.06 & 0.06 & 0.05 & 0.06 & 0.05 & 0.07 & 0.09 & 0.06 & 0.12 & 0.06 & 1.0 & 0.99 & 0.99 & 1.0 & 0.98 & 0.99 & 0.99 & 0.92 & $\textbf{1.0}$ & 0.84 & 0.98 \\
    Ours & 0.03 & 0.04 & 0.03 & 0.03 & 0.03 & 0.04 & 0.06 & $\textbf{0.03}$ & 0.11 & $\textbf{0.04}$ & 0.06 & $\textbf{1.0}$ & 0.999 & $\textbf{1.0}$ & 0.999 & 0.989 & 0.999 & 0.989 & 0.992 & 0.964 & 0.958 & 0.971 \\
    \cmidrule(lr){1-1} \cmidrule(lr){2-12} \cmidrule(lr){13-23}
    fireplace & 0 & 1 & 2 & 3 & 4 & 5 & 6 & 7 & 8 & 9 & 10 & 0 & 1 & 2 & 3 & 4 & 5 & 6 & 7 & 8 & 9 & 10 \\
    \cmidrule(lr){1-1} \cmidrule(lr){2-12} \cmidrule(lr){13-23}
    Jeon-CVPR \cite{jeon2015accurate} & $\textbf{0.05}$ & $\textbf{0.08}$ & $\textbf{0.08}$ & $\textbf{0.09}$ & $\textbf{0.1}$ & 0.09 & 0.09 & 0.1 & 0.32 & 0.09 & 0.14 & $\textbf{0.96}$ & $\textbf{0.9}$ & $\textbf{0.87}$ & 0.9 & $\textbf{0.86}$ & $\textbf{0.84}$ & $\textbf{0.85}$ & $\textbf{0.89}$ & 0.57 & 0.86 & 0.71 \\
    OACC-Net \cite{OACC-Net} & 0.08 & 0.13 & 0.14 & 0.11 & 0.13 & 0.15 & 0.14 & 0.16 & 0.35 & 0.14 & 0.15 & 0.95 & 0.86 & 0.81 & $\textbf{0.9}$ & 0.84 & $\textbf{0.84}$ & 0.85 & 0.86 & $\textbf{0.58}$ & 0.84 & $\textbf{0.8}$ \\
    Ours & 0.12 & 0.1 & 0.12 & 0.13 & 0.1 & $\textbf{0.09}$ & $\textbf{0.09}$ & $\textbf{0.09}$ & $\textbf{0.32}$ & $\textbf{0.07}$ & $\textbf{0.13}$ & 0.741 & 0.778 & 0.751 & 0.716 & 0.8 & 0.837 & 0.828 & 0.875 & 0.558 & $\textbf{0.92}$ & 0.753 \\
    \cmidrule(lr){1-1} \cmidrule(lr){2-12} \cmidrule(lr){13-23}
    living room & 0 & 1 & 2 & 3 & 4 & 5 & 6 & 7 & 8 & 9 & 10 & 0 & 1 & 2 & 3 & 4 & 5 & 6 & 7 & 8 & 9 & 10 \\
    \cmidrule(lr){1-1} \cmidrule(lr){2-12} \cmidrule(lr){13-23}
    Jeon-CVPR \cite{jeon2015accurate} & 0.04 & 0.04 & 0.05 & 0.04 & $\textbf{0.05}$ & $\textbf{0.05}$ & $\textbf{0.06}$ & 0.07 & 0.07 & 0.1 & 0.1 & 0.99 & 1.0 & $\textbf{0.99}$ & 0.99 & $\textbf{0.96}$ & 0.99 & $\textbf{0.99}$ & 0.95 & 0.99 & $\textbf{0.92}$ & 0.94 \\
    OACC-Net \cite{OACC-Net} & $\textbf{0.04}$ & $\textbf{0.04}$ & 0.05 & 0.04 & 0.08 & 0.05 & 0.07 & 0.08 & $\textbf{0.06}$ & 0.12 & 0.09 & $\textbf{1.0}$ & $\textbf{1.0}$ & 0.99 & $\textbf{0.99}$ & 0.96 & $\textbf{0.99}$ & 0.99 & $\textbf{0.95}$ & $\textbf{1.0}$ & 0.87 & $\textbf{0.96}$ \\
    Ours & 0.05 & 0.05 & $\textbf{0.04}$ & $\textbf{0.04}$ & 0.07 & 0.05 & 0.08 & $\textbf{0.06}$ & 0.12 & $\textbf{0.08}$ & $\textbf{0.07}$ & 0.974 & 0.98 & 0.983 & 0.974 & 0.93 & 0.975 & 0.92 & 0.946 & 0.928 & 0.915 & 0.878 \\
    \bottomrule
 \end{tabular}
}
\caption{Detailed reconstruction error metrics for all the scenes considering the eleven different motion scenarios (from 0 to 10) of the dataset. The upward arrow means that a higher score is better.}
\label{tab:scenes}
\end{table}

\section{Projection Model Formulation}
We provide additional details of the construction of the formulation for the RSLF projection model presented in Section 3, until Eq. (5), of the main paper. In order to describe the projection of a point in the world coordinate frame, we first apply a thin lens projection through the main lens and then a pinhole projection through every micro-lens of the micro-lens array (MLA) independently (as shown in \cref{fig:frames}).
\begin{figure}[t!]
        \centering
	    \includegraphics[height=0.4\linewidth]{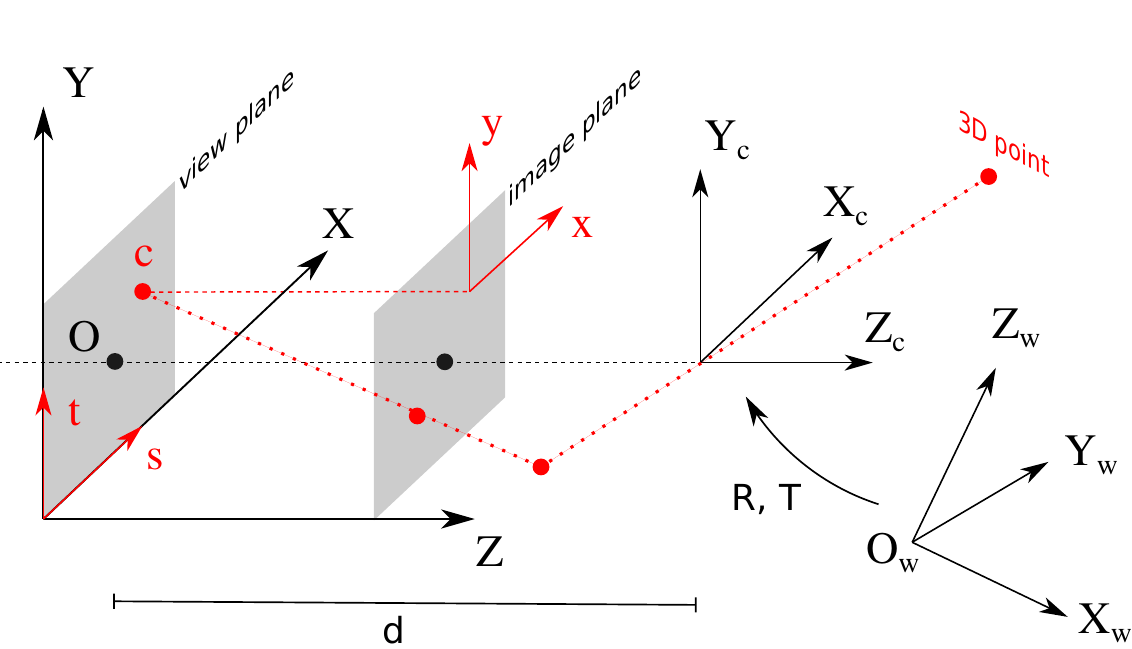}	
	    \caption{This figure is already presented in the main paper and is duplicated here for the convenience of the reader - The adopted LF coordinate frames: The 3D point is projected in a 3D virtual scene by thin lens projection, then on the 2D image plane by pinhole projection which coordinate frame depends on the considered viewpoint.}\vspace*{-0.3cm}
	    \label{fig:frames}
\end{figure}
Given a point in the world homogeneous coordinate frame ${}^w \widetilde{\mathbf{p}} =(x_w, y_w, z_w, 1)^\top$, its coordinates in the camera coordinate frame will be given by the matrix ${}^c \mathbf{M}_w {}^w$ as
\begin{equation}
\label{eq:mat_cam_pose}
{}^c \widetilde{\mathbf{p}} = {}^c \mathbf{M}_w {}^w \widetilde{\mathbf{p}},
\end{equation}

\noindent where $\widetilde{\mathbf{p}}$ represents the homogeneous coordinates of point $\mathbf{p}$ in the left-upperscripted reference frame, while ${}^c \mathbf{M}_w$ is the homogeneous matrix displacing the camera frame onto the world frame:
\begin{equation}
{}^c \mathbf{M}_w =
\left[\begin{array}{c c} 
{}^c \mathbf{R}_w & {}^c \mathbf{T}_w\\ 
\boldsymbol{0}^\top & 1 
\end{array}\right].
\end{equation}
\noindent  We will then project it through the main lens as ${}^v \widetilde{\mathbf{p}} =(x_v, y_v, z_v, 1)^\top$, using the projection matrix for a thin lens projection as
\begin{equation}
\lambda_c
{}^v \widetilde{\mathbf{p}}
=
\underbrace{
	\begin{bmatrix}
	1 & 0 & 0 & 0\\
	0 & 1 & 0 & 0\\
	0 & 0 & 1 & 0\\
	0 & 0 & -\frac{1}{F} & 1\\
	\end{bmatrix}
}_{=:\mathbf{K}_c}
{}^c \widetilde{\mathbf{p}},
\end{equation}
with $F$ the focal distance of the main lens and $\lambda_c$ a scaling factor.
To express this point $\widetilde{\mathbf{p}} =(x, y, z, 1)^\top$ in the sensor coordinate frame, we use the homogeneous matrix $\mathbf{D}$ linked to the geometry of the sensor as
\begin{equation}
\widetilde{\mathbf{p}}
=
\underbrace{
	\begin{bmatrix}
	1 & 0 & 0 & O_x\\
	0 & 1 & 0 & O_y\\
	0 & 0 & 1 & d\\
	0 & 0 & 0 & 1\\
	\end{bmatrix}
}_{=:\mathbf{D}}
{}^v \widetilde{\mathbf{p}},
\end{equation}
with $\mathbf{O}=(O_x, O_y, 0)^\top$ the intersection of the optical axis and the view plane and $d$ the distance between the optical center of the main lens and the view plane.\\
Given a point $\mathbf{c}=(s,t,0)^\top$ from the view plane, \ie a projection center, the pinhole projection to an image point $\widetilde{\mathbf{m}}^{s,t} =(x^{s,t}, y^{s,t}, 1)^\top$ is given by the matrix
\begin{equation}
\label{eq:mat_zhang}
\lambda_s^{s,t}
\widetilde{\mathbf{m}}^{s,t}
=
\underbrace{
	\begin{bmatrix}
	f & 0 & 0 & -fs\\
	0 & f & 0 & -ft\\
	0 & 0 & 1 & 0\\
	\end{bmatrix}
}_{=:\mathbf{K}_{s}^{s,t}}
\widetilde{\mathbf{p}},
\end{equation}
with $f$ the focal distance of the micro-lenses, \ie the distance between the view plane and the image plane and $\lambda_s$ a scaling factor. This is a classical pinhole projection that take into account the position of the micro-lens $\mathbf{c}$. From these equations we can find the LF point $(x,y,s,t)$ for any 3D point in the world, in the case of a LF camera modeled with the GS hypothesis. In order to add the RS effect in our model, and thus the movement of the camera in our equations, we adopt a similar RS formalism from Ait-Aider \etal~\cite{ait2006simultaneous}. They considered that the camera moves by the same little uniform movement between any two lines of pixels and  define camera pose in function of the pixel line. We also make the hypothesis that the acquisition time inside a micro-image is instantaneous (\ie micro-images are considered GS). Assuming a uniform movement $[\delta\mathbf{R}^{\delta t}\mid\delta\mathbf{T}^{\delta t}]$ between any two lines of micro-images, we can express the camera pose in function of the viewpoint line and rewrite \cref{eq:mat_cam_pose} as
\begin{equation}
\label{eq:deltaM}
{}^c \widetilde{\mathbf{p}}  =
\left[\begin{array}{c c} 
\delta\mathbf{R}^t {}^c \mathbf{R}_w & {}^c \mathbf{T}_w +\delta\mathbf{T}^t\\ 
\boldsymbol{0}^\top & 1 
\end{array}\right]
{}^w \widetilde{\mathbf{p}},
\end{equation}
with 
\begin{equation}
\label{eq:Rodrigues_1}
\delta\mathbf{R}^t = \mathbf{a}\mathbf{a}^\top (1-\cos(\Omega \tau t))+\boldsymbol{I} \cos(\Omega \tau t)+[\mathbf{a}]_\wedge \sin( \Omega \tau t), \mbox{ and  } \delta\mathbf{T}^t = \mathbf{v} \tau t ,
\end{equation}
with $\mathbf{a}$ (axis of rotation) $\Omega$ (angular velocity) and $\mathbf{v}$ (linear velocity) describes the uniform movement of the camera coordinate frame with respect to the world coordinate frame. $\tau$ is the time between the acquisition of two lines of point of view and $t$ is the line coordinate of the point of view $\mathbf{c}=(s,t,0)^\top$ from \cref{eq:mat_zhang}. The complete RSLF projection of the 3D point ${}^w \widetilde{\mathbf{p}}_i$ to a image point $\mathbf{m}_i^{s,t}$, given a center of projection $\mathbf{c}=(s,t,0)^\top$, is then
\begin{equation}
\label{eq:proj}
\lambda \mathbf{m}_i^{s,t} = \mathbf{K}_s^{s,t} \mathbf{D} \mathbf{K}_c [\delta\mathbf{R}^t {}^c \mathbf{R}_w \mid {}^c \mathbf{T}_w +\delta\mathbf{T}^t] {}^w \widetilde{\mathbf{p}}_i .
\end{equation}
That can be simplified as
\begin{equation}
\lambda \mathbf{m}_i^{s,t}  = \mathbf{K}^{s,t}[\delta\mathbf{R}^t {}^c \mathbf{R}_w \mid {}^c \mathbf{T}_w +\delta\mathbf{T}^t] {}^w \widetilde{\mathbf{p}}_i,
\end{equation}
with 
\begin{equation}
\label{eq:Kst}
\mathbf{K}^{s,t} =
\begin{bmatrix}
f & 0 & -\frac{f}{F}(O_x-s) & f(O_x-s)\\
0 & f & -\frac{f}{F}(O_y-t) & f(O_y-t)\\
0 & 0 & 1-\frac{d}{F} & d
\end{bmatrix},
\end{equation}
\noindent which is presented in Eq. (5) of the main paper. As discussed in the main paper, this projection model combines at the same time the property of a light-field sensor and a rolling shutter sensor. Specifically, the model can be extended to a GS light-field camera, when the temporal delay between two consecutive lines is zero, $\tau = 0$, and thus the position of the sensor with respect to the scene will be identical for any $t$. In fact, \cref{eq:deltaM} will be simplified into \cref{eq:mat_cam_pose}, that is the global shutter case of the light-field projection described earlier. For similar reasons, the model will act like a global shutter light-field camera when the camera has no velocity with respect to the object. The RSLF projection model generalizes to a conventional camera projection in the case where the MLA is composed of a unique lens. Indeed if we set $s = 0$ and $t= 0$, \cref{eq:Kst} becomes
\begin{equation}
\mathbf{K}^{0,0} =
\begin{bmatrix}
f & 0 & -\frac{f}{F}O_x & fO_x\\
0 & f & -\frac{f}{F}O_y & fO_y\\
0 & 0 & 1-\frac{d}{F} & d
\end{bmatrix}
\end{equation}
which correspond to a pinhole camera with projection matrix $\mathbf{K}'$ at position $\mathbf{D}'$, with
\begin{equation}
\label{eq:Dprime}
\mathbf{K}' =
\begin{bmatrix}
f & 0 & c_x\\
0 & f & c_y\\
0 & 0 & 1
\end{bmatrix},  ~~
\mathbf{D}' =
\begin{bmatrix}
1 & 0 & 0 & O_x-c_x \frac{d}{F}\\
0 & 1 & 0 & O_y-c_y \frac{d}{F}\\
0 & 0 & \frac{f O_x}{F c_x} & d
\end{bmatrix},
\end{equation}

\noindent with $c_x = (\frac{d}{F}-1) \frac{f}{F} O_x$ and $c_y = (\frac{d}{F}-1) \frac{f}{F} O_y$. Since we consider that the micro-images are locally global shutter, this pinhole model is global shutter.

\paragraph{Implementation details.} The non-linar bundle adjustment discussed in Section 3.1 of the main paper was implemented using PyTorch with the Adam optimizer, with learning rate 0.01 for 5000 iterations to ensure convergence. For the regularization of our optimization method discussed in the end of Section 3 of the main paper, we use a new coordinate frame in order to provide a center of rotation to be optimized and a normalization for the point cloud. We define the new points ${}^n \mathbf{p}_i$ as:
\begin{equation}
{}^n \mathbf{p}_i = \frac{\mathbf{p}_i - \mathbf{g}}{\lambda_n},
\end{equation}
\noindent with $\mathbf{g}$ the center of rotation and $\lambda_n$ the normalization factor.
$\mathbf{g}$ is initialized as the mean position of the initial points $\mathbf{p}_i$ and $\lambda_n$ is defined before the optimization and is calculated as 
\begin{equation}
\lambda_n = \text{max}(\mathbf{p}_i - \mathbf{g}).
\end{equation}

\end{document}